\documentclass[a4paper,fleqn]{cas-sc}

\usepackage{amssymb}
\usepackage{amsthm}
\usepackage{amsmath}
\usepackage{multirow}
\usepackage{textcomp}
\usepackage{subcaption} 
\usepackage{siunitx}
\usepackage{epstopdf}

\usepackage{graphicx}%

\usepackage{amsfonts}%

\usepackage{mathrsfs}%
\usepackage[title]{appendix}%

\usepackage{manyfoot}%
\usepackage{booktabs}%
\usepackage{algorithm}%
\usepackage{algorithmicx}%
\usepackage{algpseudocode}%
\usepackage{listings}%
\usepackage[authoryear]{natbib}
\def\tsc#1{\csdef{#1}{\textsc{\lowercase{#1}}\xspace}}
\tsc{WGM}
\tsc{QE}
\begin{document}
\let\WriteBookmarks\relax
\def\floatpagepagefraction{1}
\def\textpagefraction{.001}
\let\printorcid\relax % 可去掉页面下方的ORCID(s)

 \shorttitle{<short title of the paper for running head>}    
\shorttitle{Open set label noise learning with robust sample selection and margin-guided module}   

% \shortauthors{<short author list for running head>} 
%\shortauthors{V. {{\=A}}nand Rawat et al.}

\title[mode = title]{Open set label noise learning with robust sample selection and margin-guided module}  

\author[Zhao]{Yuandi Zhao}
\ead{dyzhao@cauc.edu.cn}
\affiliation[Zhao]{organization={College of Air Traffic Management, Civil Aviation University of China},
            postcode={300300},
            city={Tianjin},
            country={China}}

\author[Wen]{Qianxi Xia}
\ead{1746260680@qq.com}

\author[Wen]{Yang Sun}
\ead{yhgsy@shu.edu.cn}

\author[Wen]{Zhijie Wen}
\ead{wenzhijie@shu.edu.cn}
\cormark[1]
\cortext[cor1]{Corresponding author}
\affiliation[Wen]{organization={Department of Mathematics, College of Sciences,Shanghai University},
            postcode={200444},
            city={Shanghai},
            country={China}}

\author[Ma]{Liyan Ma}
\ead{wenzhijie@shu.edu.cn}
\affiliation[Ma]{organization={School of Computer Engineering and Science, Shanghai University},
            postcode={200444},
            city={Shanghai},
            country={China}}

\author[Ying1,Ying2]{Shihui Ying}
\ead{shying@shu.edu.cn}
\affiliation[Ying1]{organization={Shanghai Institute of Applied Mathematics and Mechanics, shanghai University},
            postcode={200072},
            city={Shanghai},
            country={China}}
\affiliation[Ying2]{organization={Shanghai Institute of Applied Mathematics and Mechanics, shanghai University},
            postcode={200072},
            city={Shanghai},
            country={China}}

\begin{abstract}
In recent years, the remarkable success of deep neural networks (DNNs) in computer vision is largely due to large-scale, high-quality labeled datasets. Training directly on real-world datasets with label noise may result in overfitting. The traditional method is limited to deal with closed set label noise, where noisy training data has true class labels within the known label space. However, there are some real-world datasets containing open set label noise, which means that some samples belong to an unknown class outside the known label space.
To address the open set label noise problem, we introduce a method based on Robust Sample Selection and Margin-Guided Module (RSS-MGM).
Firstly, unlike the prior clean sample selection approach, which only select a limited number of clean samples, a robust sample selection module combines small loss selection or high-confidence sample selection to obtain more clean samples.
Secondly, to efficiently distinguish open set label noise and closed set ones,  margin functions are designed to filter open-set data and closed set data.
Thirdly, different processing methods are selected for different types of samples in order to fully utilize the data's prior information and optimize the whole model.
Furthermore, extensive experimental results with noisy labeled data from benchmark datasets and real-world datasets, such as CIFAR-100N-C, CIFAR80N-O, WebFG-469, and Food101N, indicate that our approach outperforms many state-of-the-art label noise learning methods.
Especially, it can more accurately divide open set label noise samples and closed set ones.
\end{abstract}


 %We propose a RSS-MGM method, a novel algorithmic framework for learning with open-set noisy %labels. The framework makes use of the model's prediction and the unique characteristics of %OOD samples to correct ID samples and filter out OOD samples.
%Compared with traditional small-loss selection methods, our framework includes a high-%confidence sample selection method that significantly improves the model's recognition %performance for clean samples. This strategy is innovative in that it not only increases the %model's capacity to recognize open-set samples in noisy data sets, but it also optimizes the %model's overall learning process, ensuring the effective utilization of high-quality samples.
%\item Research highlight 3
%Considering the complexity and diversity of OOD samples, we design a margin-guided module. %This module employs margin functions that can accurately identify and separate OOD samples %and ID samples.  This improves the model's capacity to handle outliers and edge cases.
%\item Research highlight 4
%Experiments conducted on the standard and real-world datasets CIFAR-100N-C, CIFAR80N-O, WebFG-%496, and Food101N indicate that our method outperforms other advanced algorithms, %particularly in OLNL. Furthermore, experiments are used to verify the validity and %reasonableness of the algorithm components.

% Keywords

\maketitle

% Main text
\section{Introduction}\label{sec1}

Deep neural networks (DNNs) \citep{2012ImageNet, ren2015faster, redmon2017yolo9000, xie2020region, luo2019segeqa} have made significant advances in the domain of computer vision with large-scale and high-quality labeled datasets (e.g. ImageNet \citep{deng2009imagenet} and MSCOCO \citep{lin2014microsoft}). These datasets contain a large number of data, providing sufficient training samples for models, reducing overfitting to some extent and improving the models' generalization capabilities. On the one hand, large-scale, high-quality annotations necessitate manual annotation, which takes up a substantial amount of time and resources. \citep{li2020dividemix, wang2022promix, wu2020class2simi, sehwag2019analyzing}. On the other hand, we may easily collect low-quality image data from image search engines and other platforms \citep{sun2021webly, yang2018recognition, yao2020bridging, tanaka2018joint, yao2018extracting, zhang2019metacleaner, liu2021exploiting, sun2021co}. Label noise is unavoidable \citep{li2017webvision, xiao2015learning, sun2020crssc, yao2019towards}. Recent research shows that training with noise-labeled data would definitely influence model performance, particularly for deep neural networks with significant learning and memory abilities \citep{nettleton2010study, zhang2021understanding, arpit2017closer}. Therefore, learning with noisy label (LNL) is essential to preserve model reliability and flexibility in real-world applications.

\begin{figure}
	\centering
	\includegraphics[width=0.7\linewidth]{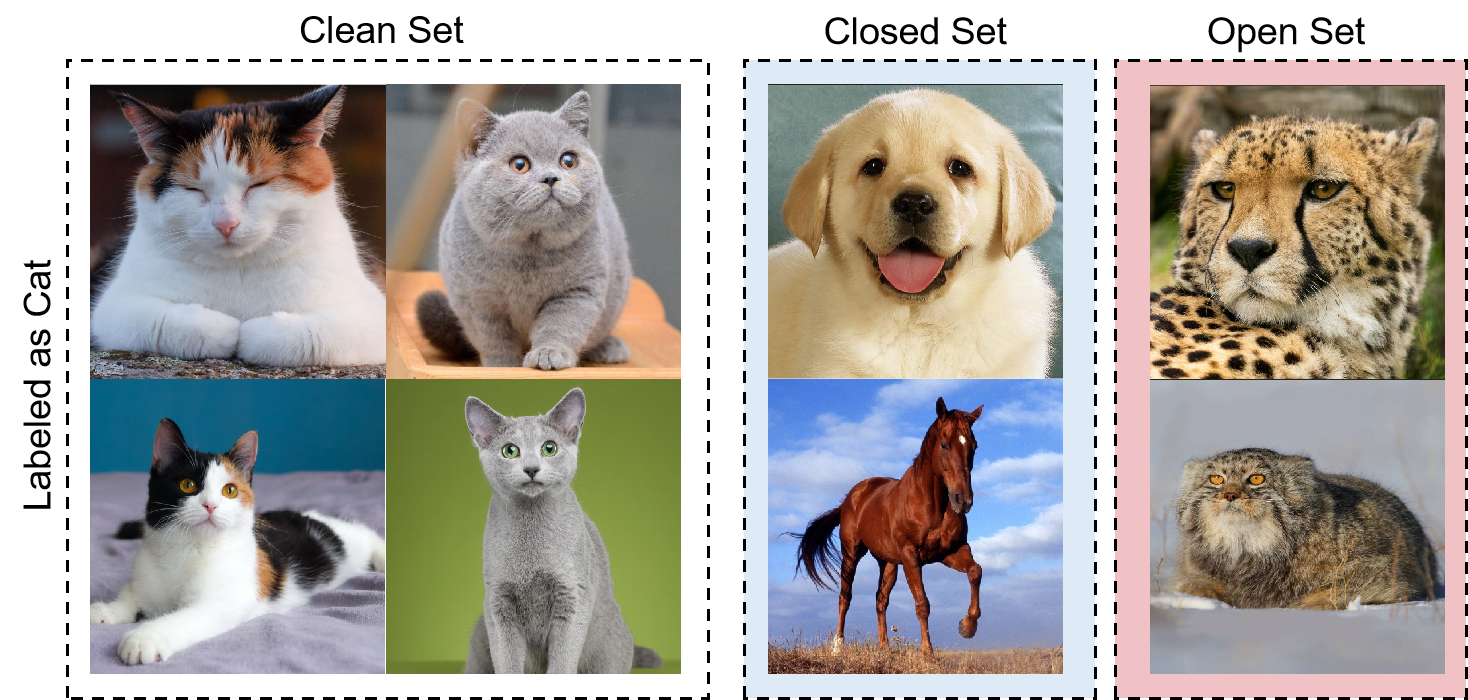}
	\caption{The example of open set label noise problem. Collecting a network dataset with labels from the CIFAR10 dataset but different images. Each image in the dataset is divided into three different groups: Clean Set, Closed Set, and Open Set. Clean Set refers to images with correct labels. Closed Set signifies images that are labeled incorrectly, yet their correct labels still exist within the known label space. Open Set indicates that the image is incorrectly labeled and the ground-truth label is outside the label space.}
	\label{fig:figure1}
\end{figure}

In real-world datasets, noisy labels can be classified into two types: closed-set noise and open-set noise. \citep{wang2018iterative}.
As shown in Fig. \ref{fig:figure1}, we use a search engine to collect a web dataset containing images labeled using the CIFAR-10 dataset but with different images. This dataset can be separated into three categories: a clean set, a closed set, and an open set.
For example, the sample labeled "cat", the white background box represents the clean set, includes images that are actually cats.
The blue background box represents the closed set, which includes images that are not cats but belong to other classes within the label space, such as dog or horse.
And the red background box represents the open set, which includes images that are not cats and don't belong to any other classes within the label space, such as leopard or pallas cat.
Closed set noise indicates that the true label of the noise sample is within the known label space of the training dataset $\mathcal{Y}_{known}$. In other words, the true class of the closed set noise sample falls within the category of classes which we are already familiar.  In contrast, open set noise arises when the correct label for a noisy example lies outside the label space $\mathcal{Y}_{known}$ defined by the training dataset, which means these samples belong to an unknown class not seen by the training set. In summary, closed-set noise samples are considered to be part of the data's in-distribution (ID), whereas open-set noise samples are part of the data's out-of-distribution (OOD).

In LNL, the majority of research primarily addresses the challenge of closed set label noise, such as loss correction \citep{lin2017focal, ghosh2017robust, zhang2018generalized}, label correction \citep{wu2018light, reed2014training, arazo2019unsupervised, zheng2020error}, and sample selection \citep{zhang2017mixup, han2018co, lee2018cleannet}. However, these solutions are frequently constrained when faced with practical application problems. Meanwhile, open set noisy labels often appear in real-world datasets and tend to present a more intricate challenge. Despite this, several studies already explores this issue, such as ILON \citep{wang2018iterative}, which selects clean samples using an iterative learning framework, has begun looking into the open-set noisy labels problem. In addition, InsCorr \citep{xia2021instance} uses partially discarded data information for semi-supervised learning. However, these methods frequently assume that the training dataset is only impacted by closed set noise or open set ones, without taking into account the possibility of both. EvidentialMix \citep{sachdeva2021evidentialmix} and MoPro \citep{li2020mopro} propose that noise labels include both closed set noise or open set ones, but they don't use the information from the discarded data. Jo-SRC \citep{yao2021jo} and PNP \citep{sun2022pnp} attempt to identify mislabeled open set data, whereas Co-mining \citep{cai2023co} seeks to differentiate between hard samples and noisy samples by creating discriminative and robust feature spaces.
However, due to the unavailability and diversity of open set samples, the model has difficulty identifying them. In some cases, the model is unable to successfully distinguish open set samples from clean data.

In this paper, we propose a novel deep learning method called Robust Sample Selection and Margin-Guided Module (RSS-MGM), which aims to address the challenges posed by closed-set noise and open-set noise on image classification tasks. To address these issues, we focus on sample selection methods that enhance the robustness and reliability of the model by combining robust sample selection and margin-guide learning.

First the robust sample selection module combines a small loss selection or high confidence sample selection, which can select more clean samples. The Jensen-Shannon (JS) divergence is selected to measure the similarity between predicted probabilities and true labels to identify clean samples. Then, we calculate the confidence scores of the samples to further filter the clean samples and construct a bigger clean sample subset.
Second, we introduce the margin-guided module, which distinguishes between ID and OOD samples. We propose margin functions to choose open set samples and closed set ones based on their respective properties. Then, the training dataset is divided into three sets: clean set, closed set, and open set.
Finally, to address different types of samples, we employ a customized loss function. For clean samples, we use a cross-entropy loss function. For ID noise samples, we use a specific loss function and introduce a consistency regularization loss to improve the consistency and generalization ability of the model.

The main contributions of this paper are summarized as follows:

1. We propose a RSS-MGM method, a novel algorithmic framework for learning with open-set noisy labels. The framework makes use of the model's prediction and the unique characteristics of OOD samples to correct ID samples and filter out OOD samples.

2. Compared with traditional small-loss selection methods, our framework includes a high-confidence sample selection method that significantly improves the model's recognition performance for clean samples. This strategy is innovative in that it not only increases the model's capacity to recognize open-set samples in noisy data sets, but it also optimizes the model's overall learning process, ensuring the effective utilization of high-quality samples.

3. Considering the complexity and diversity of OOD samples, we design a margin-guided module. This module employs margin functions that can accurately identify and separate OOD samples and ID samples.  This improves the model's capacity to handle outliers and edge cases.

4. Experiments conducted on the standard and real-world datasets CIFAR-100N-C, CIFAR80N-O, WebFG-496, and Food101N indicate that our method outperforms other advanced algorithms, particularly in OLNL. Furthermore, experiments are used to verify the validity and reasonableness of the algorithm components.

The remainder of the paper is structured as follows. In Section \ref{sec2}, we provide an overview of the background on learning with open-set label noise. In Section \ref{sec3}, we describe the RSS-MGM method in detail. Experimental results are shown in Section \ref{sec4}. Finally, the paper is summarized in Section \ref{sec5}.

\section{Related Works}\label{sec2}

\paragraph{Learning With Noisy Label}
Several methods have been proposed for LNL, such as loss correction \citep{lin2017focal, ghosh2017robust, zhang2018generalized}, label correction \citep{wang2022promix, wu2018light, reed2014training, arazo2019unsupervised, zheng2020error, albert2023your, xu2023label}, and sample selection \citep{zhang2017mixup, han2018co, lee2018cleannet, yu2019does, pleiss2020identifying, cheng2020learning, zhu2021hard, zhang2022model, cordeiro2023longremix,  wen2023jsmix, li2024103801}.

Loss Correction aims to adjust the loss function to mitigate the influence of noise labels while training. For instance, Lin et al. \citep{lin2017focal} develop the Focal Loss, which use an adjustable parameter to reduce the influence of noise labels on easily classified samples. 
Ghosh et al. \citep{ghosh2017robust} propose the Mean Absolute Error (MAE) Loss, which use absolute error in evaluating the difference between predicted labels and true labels. 
Conversely, the Cross-Entropy (CE) Loss emphasizes challenging samples and reduces the influence of noisy samples by modifying the weights of the loss function.
In addition, Zhang et al. \citep{zhang2018generalized} generalize CE Loss to GCE Loss, which includes an adjustable parameter to alter the weighting for different classes, increasing the model's robustness to noisy samples.

Label correction aims to improve the quality of training data by estimating and correcting noisy labels.
Wu et al. \citep{wu2018light} propose using a similar neighborhood of clean data to train a classifier and assign pseudo-labels to noisy samples. 
Reed et al. \citep{reed2014training} propose the Bootstrapping technique to handle noisy labels. 
Arazo et al. \citep{arazo2019unsupervised} propose an unsupervised label noise modeling and loss correction, which automatically learns the probability distribution of noisy labels and adjusts the loss function. 
Zheng et al. \citep{zheng2020error} propose an error-constrained label correction method to limit the range of label correction.

Sample selection efficiently filter training samples in order to reduce the interference of noisy labels.
Zhang et al. \citep{zhang2017mixup} propose a sample mixing strategy (Mixup) that linearly estimates features from several samples, generating new samples for training. 
Han et al. \citep{han2018co} propose Co-teaching, where two models collaborate by selecting samples to train each other. 
Lee et al. \citep{lee2018cleannet} propose CleanNet, which uses feature vector similarity to choose clean samples through pruning.
Li et al. \citep{li2024103801} proposed the Twin Binary Classification-Mixed Input (TBC-MI) method, which uses a twin binary classification network to transform complex multiclass classification problems into simpler binary classification tasks.

\paragraph{Open Set Label Noise Learning}

Open set label noise learning (OLNL) attempts to developing more robust and reliable deep learning models by learning from real-world datasets that contain both open set noise and closed set one. 
A variety of related works have been proposed.
For instance, Wang et al. \citep{wang2018iterative} design an iterative learning framework that enables effective robust training on datasets with open set label noise. 
Sachdeva et al. \citep{sachdeva2021evidentialmix} specifically design the EvidentialMix framework to capture and analyze the impact of both open set and closed set noise.
Xia et al. \citep{xia2021instance} employ instance correction techniques to adjust and optimize the instances of discarded data. 
Li et al. \citep{li2020mopro} introduce Momentum Prototypes (MoPro), which effectively corrects ID samples and drops OOD samples. 
Yao et al. \citep{yao2021jo} propose a consistency-based method to identify open set noise. 
Sun et al. \citep{sun2022pnp} achieve more accurate sample selection by modeling the consistency of different predictions for the same input sample. 
Cai et al. \citep{cai2023co} focus on constructing a feature space to effectively distinguish between hard samples and noisy samples.
Wan et al. \citep{wan2024unlocking} proposed Class Expansion Contrastive Learning (CECL), an innovative two-step contrastive learning framework designed to effectively handle open set label noise.
\paragraph{Contrastive Learning}

Contrastive learning \citep{chen2020simple, chen2020improved, ortego2021multi} has grown in popularity in recent years due to its capacity to get key information from unlabeled data. This approach is making significant progress in a number of applications \citep{yang2021taco, li2021learning}. 
Specifically, each input image is randomly augmented, resulting in two augmented views of the image. Subsequently, the model is optimized using a contrastive loss function. This strategy enhances the model's capacity to differentiate image features, increasing the accuracy and robustness of the model.
For instance, SimCLR \citep{chen2020simple} is a simple yet effective contrastive learning framework that emphasizes the importance of data augmentation. 
MoCo \citep{chen2020improved} establishs the concept of momentum contrast, which enhances stability by maintaining momentum updates in the network. 
PCL \citep{jiang2020beyond} uses prototypes as cluster centers for contrastive learning in the primary subspace, which further enhancements utilizing weakly supervised contrastive losses and Mixup \citep{zhang2017mixup}. 
RRL \citep{li2021learning} aims to enhance model resilience to noisy data by adversarially addressing noise-induced interference. 
LaCoL \citep{yan2022noise} emphasizes the latent value of noise through negative correlation, aiming to mine information from noisy samples to enrich learning representations.

\section{The Proposed Method: RSS-MGM}\label{sec3}

We define the training dataset as $\mathcal{D}(x_i,y_i)_{i=1}^{N}$, where $x_i$ is an input image and $y_i=\mathcal{C} \in \{1,...,C\}$ is its annotated label. Denote $y_i^\star$ as the ground-truth label of $x_i$. 

Traditionally, it has been assumed that all annotated labels are correct (i.e., $y_i = y_i^\star$), allowing the model to be optimized by minimizing experience loss.:
\begin{equation}
	\mathcal{L}=\mathbb{E}_{\mathcal{D}}[l_{ce}(x_i,y_i)]=\frac{1}{N}l_{ce}(x_i,y_i),
\end{equation}
in which,
\begin{equation}
	l_{ce}(x_i,y_i)=-\sum_{c=1}^{C}y_i^clog(p^c(x_i,\Theta)),
\end{equation}
where $\Theta$ represents the model parameters, and $p^c(x_i,\Theta)$ represents the network's prediction that the sample $x_i$ belongs to class $c$. For convenience, we represent $p^c(x_i, \Theta)$ as $p_i^c$. $y_i^c$ indicates the class label for sample $x_i$ in class $c$.

However, when a dataset contains noise labels (e.g., a web image dataset), the assumption that the labels are completely clean is incorrect. This is due to the difficulty of identifying the complexity of big and different datasets, which inevitably leads to labeling errors. Despite the high model capacity of DNNs, they often overfit to noisy labels because of the network's memorizing effect, severely impacting the model's classification accuracy and generalization ability \citep{DBLP:conf/iclr/ZhangBHRV17}.

\begin{figure}
	\centering
	\includegraphics[width=1\linewidth]{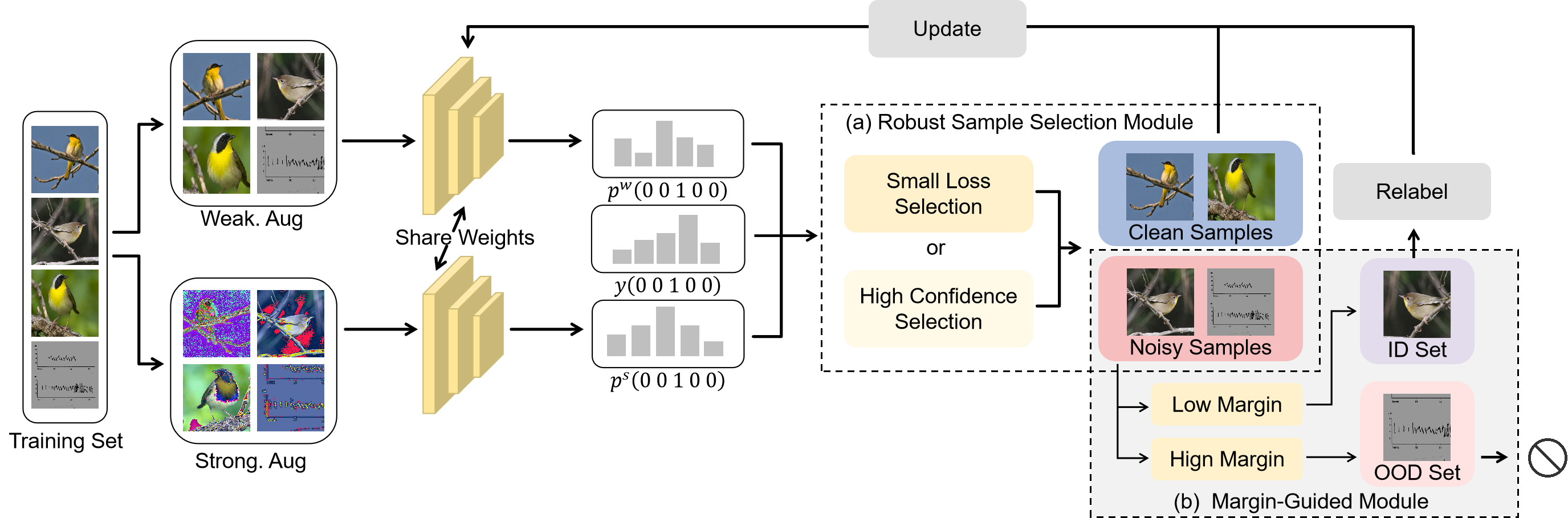}
	\caption{The overall framework of RSS-MGM. Each input image $x_i$ is augmented into one weakly view and one strongly augmented view before being fed into the label predictor network, leading to two label predictions: $p^w$ for the weakly augmented view and $p^s$ for the strongly augmented view. Afterward, based on the Robust Sample Selection Module, samples are classified as Clean Set or Noisy Set. If the sample is clean, it will be fed into the label prediction network. Otherwise, based on the Margin-Guided Module, samples are divided into ID Set or OOD Set. Samples from the ID Set will be re-labeled to update the network, while samples from the OOD Set will be directly discarded. Finally, our model is updated by back-propagating.}
	\label{fig:framework}
\end{figure}

The aim of this research is to train the network on a dataset with label noise, including both open set noise and closed set one, with the goal of obtaining high accuracy on test set. By addressing the challenges given by these noisy labeled data, we hope to improve the network's robustness and classification performance in a wide range of real-world situations. Fig. \ref{fig:framework} illustrates the algorithm's general logic and training framework.
Specifically, the training dataset is augmented to generate a weakly augmented view and a strongly augmented view. These views are then fed into two separate networks with shared weights for prediction. The prediction results $p^s$ and $p^w$, as well as the label $y$, are fed into (a) the robust sample selection module, which divides the training set into the clean set and the noise set. Then, the noise set is fed into (b) the margin-guided module, which further divides the noise set into ID set and OOD set. Samples from the the clean set and the ID sets are used to update the network, while samples from the OOD set are discarded.

\subsection{Robust Sample Selection Module}

When noisy labels are presence in the dataset, there is a possibility that the model will learn their incorrect relationships, causing a performance drop.
Previous studies typically employ small-loss sample selection methods to choose clean samples. However, this approach often results in a small number of clean samples. This limitation becomes particularly problematic in OLNL, where continuing to use this method may exacerbate the performance decline of the model.
To address this issue, we propose a robust sample selection module that filters bigger clean samples for network training. 

We use the loss function to select samples and divide the training dataset into clean samples and noisy ones. DNNs tend to prioritize learning clean samples initially and progressively adapting to the inclusion of noisy samples \citep{li2017webvision, zhang2021understanding}.
We select clean samples based on Jensen-Shannon (JS) divergence. For a sample $x_i$, we utilize JS divergence to measure the dissimilarity between the predicted  distribution $p_i=[p_i^1,p_i^2,...,p_i^C]$  and its corresponding label distribution $y_i=[y_i^1,y_i^2,...,y_i^C]$. 
\begin{equation}
	d_i = D_{JS}(p_i||y_i)=\frac{1}{2}D_{KL}(p_i||\frac{p_i+y_i}{2})+\frac{1}{2}D_{KL}(y_i||\frac{p_i+y_i}{2}),
	\label{eq:d_i}
\end{equation}
in which
\begin{equation}\label{js-div}
	D_{KL}(p_i||p_j) = \sum_{c=1}^{C}p^c_i\log\frac{p^c_i}{p^c_j},
\end{equation}
where \( D_{KL}(p_i || p_j) \) is the Kullback-Leibler divergence, which measures the difference between two probability distributions.

A lower JS divergence indicates a higher similarity between the predicted distribution of the sample and its corresponding label distribution. Conversely, a higher JS divergence suggests a greater difference between them, potentially containing noise. Therefore, we can use JS divergence to select clean samples.

When considering the logarithm with base 2, the JS divergence is limited to 0 and 1. This can be regarded as a probability that the sample \(x_i\) is a clean sample.

Then we define a $\mathcal{P}$ as follows:
\begin{equation}
	\mathcal{P} = 1-d_i.
\end{equation}

After introducing a threshold \(\tau_s \), we filter out samples with probabilities below this threshold, resulting in our defined "clean" set,
\begin{equation}
	\mathcal{D}_{c_s} = \{ x_i|\mathcal{P}_i > \tau_s \}.
	\label{eq:fliter-1}
\end{equation}
where $\mathcal{D}_{c_s}$ refers to the clean sample set after small-loss selection.

Then, the "unclean" set can be represented as $\mathcal{D}_n = \mathcal{D} \backslash \mathcal{D}_{c_s}$. However, this selection strategy does not filter out all clean samples, making it unsuitable for OLNL.
As shown in Figure \ref{fig:small-loss-fewer-samples}, Clean Set and Noise Set represent the sets obtained by small-loss selection, respectively. The samples bordered in red in the noise set are clean, but are incorrectly selected into the noise set.

Inspired by the pseudo-labeling method used for semi-supervised learning \citep{sohn2020fixmatch, zhao2022lassl,wang2020probabilistic,gu2016robust}, we create a larger clean set, ensuring its quality.
Specifically, we define the confidence score for any given sample $x_i$ as:

\begin{figure}
	\centering
	\includegraphics[width=0.7\linewidth]{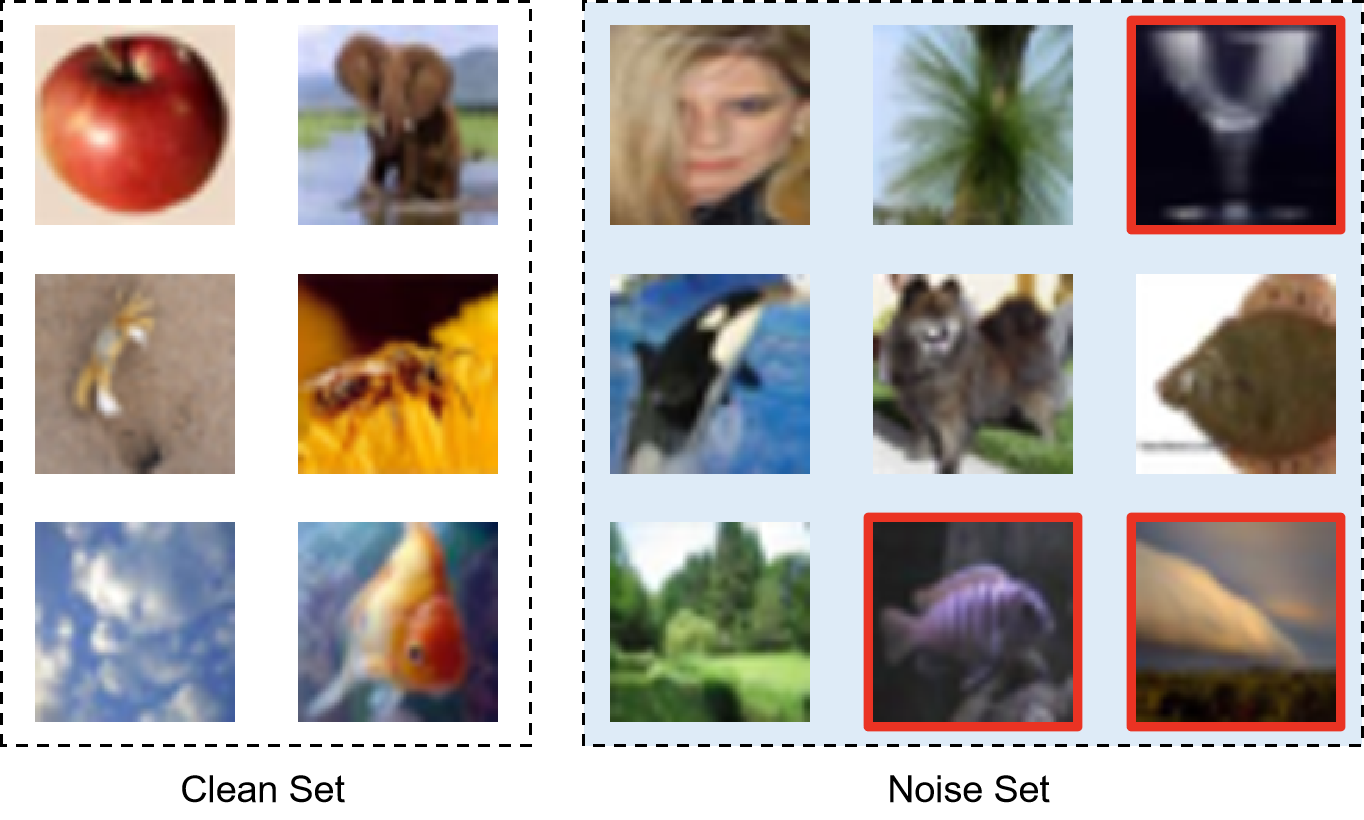}
	\caption{An example of the small loss selection method for dividing the training dataset. Here, Clean Set represents the set of samples that are identified as clean samples, which are considered to be labeled accurately. The Noise Set, on the other hand, represents the set of samples with noisy labels. However, the samples bordered in red in the noise set are actually clean, but are incorrectly classified in the noise set.}
	\label{fig:small-loss-fewer-samples}
\end{figure}
\begin{equation}
	s_i = \sum_{j=1}^C f_j(x_i)\mathbb{I} (j=y_i),
	\label{eq:score}
\end{equation}
where $f_j(x_i)$ represents the predicted probability of sample $x_i$ belonging to class $j$, and the indicator function $\mathbb{I}(j=y_i)$ returns 1 when $j$ matches the true label $y_i$ and 0 otherwise.

Then, the bigger clean set can be calculated as follows:
\begin{equation}
	\mathcal{D}_{c_h} = \{ x_i|s_i > \tau_h \},
	\label{eq:fliter-2}
\end{equation}
where $\mathcal{D}_{c_h}$ refers to the clean sample set after high confidence sample selection, \(\tau_h \) is a discriminative threshold to divide samples into two groups based on their confidence scores \(s_i \): confident samples and unconfident ones.

Finally, we can combine $\mathcal{D}_{c_s}$ and $\mathcal{D}_{c_h}$ to obtain a complete clean dataset $\mathcal{D}_c$:
\begin{equation}
	\mathcal{D}_{c} = \mathcal{D}_{c_s} \cup \mathcal{D}_{c_h}.
	\label{eq:D_clean}
\end{equation}

The noisy set $\mathcal{D}_n$ is calculated as follows:
\begin{equation}
	\mathcal{D}_{n} = \mathcal{D} \backslash \mathcal{D}_{c}.
	\label{eq:D_unclean}
\end{equation}

\subsection{ Margin-Guided Module}

In the previous section, we utilize the robust sample selection module to separate the training set $\mathcal{D}_{train}$ into $\mathcal{D}_{c}$ and $\mathcal{D}_{n}$. Despite the presence of label noise in the data from $\mathcal{D}_{n}$, it is vital to keep some samples and explore the possibility of re-labeling, particularly high confidence samples within the ID samples, as they can provide beneficial information for our model.\citep{zhang2022improving} Therefore, simply removing unclean samples is not a viable option. We need to distinguish the high confidence samples from the noisy samples and use them to improve the training of the model. 

In DNNs, the model's predictions are usually limited by the known label space. As a result, when dealing with OOD samples, the model's predictions frequently display higher label uncertainty, resulting in a nearly uniform distribution of predicted probabilities, and predicted probabilities are closer to each other. In contrast, for ID samples, the predicted probability distribution resembles a one-hot distribution, with one probability reaching the highest value. In such cases, we can select samples based on the distribution of the predicted probabilities.

To fully utilize the information, we propose to divide $\mathcal{D}_{n}$ into $\mathcal{D}_{OOD}$ and $\mathcal{D}_{ID}$, where $\mathcal{D}_{ID}$ requires further subdivision into $\mathcal{D}_{high}$ and $\mathcal{D}_{rest}$. Next, we can discard OOD and rest samples directly, while for high confidence samples, we can use semi-supervised learning to make use of their information. To achieve this purpose, we present the Margin-Guided Module.

For OOD samples, the model's predictions have a greater label uncertainty, and the probability distribution is nearly uniform. As a result, the model's predictions barely different between two alternative augmentations of the same sample $x_i$. We develope a margin ranking function to filter OOD samples based on this property.

Specifically, We use two different augmentation approaches, weak augmentation $A^w(\cdot)$ and strong augmentation $A^s(\cdot)$, to create two alternative augmented views (e.g., $v^w = A^w(x_i)$ and $v^s = A^s(x_i)$) from a given sample $x_i$. Subsequently, these views are separately fed into the network to obtain prediction probabilities, $p^w$ and $p^s$, respectively. Finally, we calculate the margin between these two views as:
\begin{equation}
	M_v = \mathbb{I}[ \text{argmax}_c p^w \ne \text{argmax}_c p^s ].
	\label{eq:margin}
\end{equation}

Based on \(M_v\), we can obtain \(\mathcal{D}_{OOD}\):
\begin{equation}
	\mathcal{D}_{OOD} = \{ x_i  | M_v^i \cdot \mathcal{D}_n, x_i \in \mathcal{D}_n \}.
	\label{eq:D_ood}
\end{equation}

And the ID set can be obtained through:
\begin{equation}
	\mathcal{D}_{ID} = \mathcal{D}_{n} \backslash \mathcal{D}_{OOD}.
	\label{eq:D_id}
\end{equation}

Thus, we have divided \(\mathcal{D}_{n}\) into \(\mathcal{D}_{OOD}\) and \(\mathcal{D}_{ID}\).

After that, we process \(\mathcal{D}_{ID}\) further. The differences in the predicted probabilities for sample \(x_i\) are remarkable since the network's predicted probability for ID samples has a maximum. We design a margin ranking function to eliminate high confidence samples in order to take advantage of these feature.

Specifically, for a given sample $x_i$, we calculate the difference between the largest value and second largest value of the model output. This measures the difference between the prediction class of the network and other classes. The margin function is defined as follows:
\begin{equation}
	M = z_y(x_i) - max_{k \ne y}z_k(x_i),
\end{equation}
where $z_y(x_i)$ represents the logit of the network's prediction for sample $x_i$ with class $y$ , $max_{k \ne y}z_k(x_i)$ is the maximum logit value of the network's prediction for sample $x_i$ among all classes except class $y$. In other words, it is the highest logit value for all classes other than class $y$ in the network's prediction for sample $x_i$.

The higher the margin value, the higher confidence of the model, enabling us to utilize the information from these samples for semi-supervised learning.
On the contrary, lower margin values indicate lower confidence in the model's predictions, suggesting potential uncertainty. This uncertainty may lead to a decrease in the model's classification performance. Therefore, we are less likely to employ the information obtained from these samples.

To ensure fairness, we consider two different views of the same sample (e.g., $v^w$ and $v^s$). 
\begin{equation}
	M_{p} = \frac{1}{2} (M_{p^w} + M_{p^s}),
	\label{eq:M_p}
\end{equation}
where
\begin{equation}
	M_{p^w} = z_y(v^w) - max_{k \ne y}z_k(v^w),
\end{equation}
\begin{equation}
	M_{p^s} = z_y(v^s) - max_{k \ne y}z_k(v^s).
\end{equation}

Consequently, given $\tau_p$, \(\mathcal{D}_{high}\) is defined as follows:
\begin{equation}
	\mathcal{D}_{high} = \{x_i | M_p^i > \tau_p \text{ and } x_i \in \mathcal{D}_{ID} \}.
	\label{eq:D_hard}
\end{equation}

The remaining samples are indicated as \(\mathcal{D}_{rest}\):
\begin{equation}
	\mathcal{D}_{rest} = \mathcal{D}_{ID} \backslash \mathcal{D}_{high}.
	\label{eq:D_rest}
\end{equation}

This way, we successfully partition \(\mathcal{D}_{n}\) into \(\mathcal{D}_{OOD}\) and \(\mathcal{D}_{ID}\), with \(\mathcal{D}_{ID}\) further divided into \(\mathcal{D}_{high}\) and \(\mathcal{D}_{rest}\).

\subsection{Label Reassignment}

In the first two parts, we successfully divide the training dataset $\mathcal{D}$ into  $\mathcal{D}_{c}$, $\mathcal{D}_{OOD}$, and $\mathcal{D}_{ID}$. Next, we apply different processing strategies on these three subdatasets.

For $\mathcal{D}_{clean}$, we assump that all sample labels are correct, thus we keep the sample labels. To improve generalization performance, we use label smoothing regularization (LSR) \citep{yao2021jo} to calculate the loss of clean samples, which transform their labels $y_i^c \in \{1,...,C\}$ to $\tilde{y}_i^c$.
\begin{equation}
	\tilde{y}_i^c=\left\{\begin{array}{cc}
		1-\epsilon, & c=y_i^c \\
		\frac{\epsilon}{C-1}, & c \neq y_i^c
	\end{array},\right.
	\label{eq:clean label}
\end{equation}
where $\epsilon$ is a hyperparameter that controls the smoothness of the label distribution.

For \(\mathcal{D}_{ID}\), we believe that \(\mathcal{D}_{high}\) can provide valuable information, therefore, we adopt a re-labeling strategy.
Specifically, we use temperature-based sharpening method to deal with these samples. We introduce a temperature parameter $\tau$ to modify the original network prediction probabilities $p_i$, which generates sharpened predictions $\hat{p}_i$.
\begin{equation}
	\hat{p}_i = \frac{{p_i}^{\frac{1}{\tau}}}{{\sum_{j} {p_j}^{\frac{1}{\tau}}}},
	\label{eq:p_i}
\end{equation}
where $p_i$ represents the original prediction of sample $x_i$, and $\tau$ is the temperature parameter used for sharpening. 

Consider the model's consistency in predicting different augmented views of sample $x_{id}$ within the same distribution. We obtain pseudo-labels for each sample by calculating the average probabilities of two different views. Therefore, the calculation of the same sample $x_i $ forecast the probability of two different augmentation $p_{x_i} ^ {w} $ and $p_{x_i} ^ {s} $, and take its averaging $\bar{p}_i $,
  \begin{equation}
	\bar{p}_i = \frac{1}{2} (p_{x_i}^{w} + p_{x_i}^{s}).
	\label{eq:hard label}
\end{equation}

Specifically for $\mathcal{D}_{rest}$, we consider that $\mathcal{D}_{rest}$ samples provide limited useful information and can be disregarded.

For $\mathcal{D}_{ood}$,  we remove these samples from the training set. There are two primary reasons. First, OOD samples may introduce unexpected changes and perturbations, affecting the model's generalization ability. Second, the labels of OOD samples do not fall inside the known label space, which may provide additional uncertainty in the model's predictions of OOD data.

\subsection{The Loss Function}
When dealing with different subsets, we apply different loss functions for optimizing. For $\mathcal{D}_{c}$, we utilize the traditional cross entropy loss function, which is defined as follows:
\begin{equation}
	\mathcal{L}_{c} = -\sum_{i \in \mathcal{D}_{c}}(\sum_{c=1}^{C} y_i^c \log p_i^c - \sum_{c=1}^{C} p_i^c \log (p_i^c)),
\end{equation}
where $C$ denotes the number of classes, $y_i^c$ is the ground-truth label of sample $x_i$, $p_i^c$ is the model's predicted probability for sample $x_i$ belonging to class $c$, and $\mathbb{I}(\cdot)$ is the indicator function. 

For $\mathcal{D}_{high}$, we propose an improved cross-entropy loss function as follows:
\begin{equation}
	\mathcal{L}_{n} = -\sum_{i \in \mathcal{D}_{id}} \sum_{c=1}^{C} y_i^c \log \left(\frac{1}{2}p_{i}^w + \frac{1}{2}p_{i}^s \right) -  \lambda \sum_{c=1}^{C} p_i^c \log (p_i^c),
\end{equation}
where $C$ denotes the number of classes, $y_i^c$ is the ground-truth lable of sample $x_i$, \(p_{i}^w\) represents the model's prediction for the weak augmented view \(x_i^w\) of the sample \(x_i\) , \(p_{i}^s\) represents the model's prediction for the strong augmented view \(x_i^s\) of the sample \(x_i\), $\mathbb{I}(\cdot)$ is the indicator function and $\lambda$ is the hyperparameter controlling the weight of the entropy regularization term.

We introduce a consistency regularization loss $\mathcal{L}_{cons}$ to encourage well-trained models to maintain consistency when predicting different augmentations of the sample \(x_i\).
\begin{equation}
	\mathcal{L}_{cons} = \frac{1}{2} \sum_{i \in \mathcal{D}} \left[ KL(p_{i}^w \parallel p_{i}^s) + KL(p_{i}^s \parallel p_{i}^w) \right],
\end{equation}
where $\mathcal{D}$ represents the training set, $p_{i}^w$ and $p_{i}^s$ is the predicted distributions for sample $x_i$ from weak and strong augmented views, respectively. $KL(\cdot \parallel\cdot)$ represents the Kullback-Leibler (KL) divergence.

The overall loss function can be represented as:
\begin{equation}
	\mathcal{L}_{total} = \mathcal{L}_{c} + \lambda_1 \mathcal{L}_{n} + \lambda_2 \mathcal{L}_{cons},
	\label{eq:L}
\end{equation}
where $\lambda_1$ and $\lambda_2$ are hyper-parameters.

\begin{algorithm}
	\caption{RSS-MGS Algorithm}\label{alg:train}
	\begin{algorithmic}[1]
		\Require Training set $\mathcal{D}=\{(x^{(n)},y^{(n)})_{n=1}^N\}$, network $f$ with parameters $\Theta$, learning rate $\eta$, epoch $T_{\text{max}}$ , iterations $I_{\text{max}}$
		\Ensure Updated network $f$
		
		\For{$t = 1$ to $T_{max}$}
		\State Shuffle training set $\mathcal{D}$;
		
		\For{$n = 1$ to $I_{max}$}
		\State Fetch mini-batch $\mathbb{B}$ from $\mathcal{D}$;
		\State Calculate $p_i = f(x_i,\Theta)$ for a sample $x_i$;
		
		\State Calculate $d_i$ and $s_i$ by Eq. (\ref{eq:d_i}) and Eq. (\ref{eq:score});
		\State Select $\mathbb{B}_{c_s}$ and $\mathbb{B}_{c_h}$ of $\mathbb{B}$ according to Eq. (\ref{eq:fliter-1}) and Eq. (\ref{eq:fliter-2});

		\State Obtain the clean set $\mathbb{B}_{c}$ by Eq. (\ref{eq:D_clean}) and the unclean set $\mathbb{B}_{n}$ by Eq. (\ref{eq:D_unclean});
		
		\If{sample $x_i$ in $\mathbb{B}_{c}$}
		\State Label smoothing regularization to $x_i$ by Eq. (\ref{eq:clean label});
		\Else
		\State Generate two different views $v^w=A^w(x_i)$ and $v^s=A^s(x_i)$;
		\State Calculate predicted probabilities $p^w$ and $p^s$ for $v^w$ and $v^s$;
		\State Calculate margin function $M_v$  and $M_p$ by Eq. (\ref{eq:margin}) and Eq. (\ref{eq:M_p});
		\State Obtain $\mathbb{B}_{OOD}$ and $\mathbb{B}_{ID}$ by Eq. (\ref{eq:D_ood}) and Eq. (\ref{eq:D_id});
		\State Divide $\mathbb{B}_{ID}$ into $\mathbb{B}_{high}$ and $\mathbb{B}_{rest}$ by Eq. (\ref{eq:D_hard}) and Eq. (\ref{eq:D_rest});
		\For{sample $x_i$ in $\mathbb{B}_{high}$}
		\State Sharpen predicted probabilities $\hat{p}_i$ by Eq. (\ref{eq:p_i});
		\State Calculate pesudo-labels for $x_i$ by Eq. (\ref{eq:hard label});
		\EndFor
		
		\EndIf

		\State Calculate loss $\mathcal{L}_{total}$ by Eq. (\ref{eq:L});
		\State Update $\Theta = \Theta - \eta \nabla L$.

		\EndFor
		\EndFor

	\end{algorithmic}
\end{algorithm}

\section{Experiments}\label{sec4}
In this section, we report the results of the RSS-MGM method and compare it with some state-of-the-art algorithms. Specifically, we evaluate the effectiveness of the proposed method on four benchmark datasets: CIFAR100N-C, CIFAR80N-O \citep{krizhevsky2009learning}, WebFG-496 \citep{sun2021webly}, and Food101N \citep{bossard2014food}. Finally, ablation studies demonstrate the effectiveness of RSS-MGM.

\subsection{Datasets and experimental settings}
\paragraph{CIFAR100N-C and CIFAR80N-O}
CIFAR100N-C and CIFAR80N-O are synthetic noise datasets created based on the CIFAR100 dataset using predefined criterion \citep{krizhevsky2009learning}. The original CIFAR100 dataset consists of 100 classes, with each containing 600 images of size 32 $\times$ 32. There are 500 training images and 100 testing images per class. Each image comes with two labels: coarse-labels and fine-labels, corresponding to the classes and superclass . Both the training set and validation sets are affected by label noise. The predefined criterion for generating synthetic noise data sets are shown as follows:

CIFAR100N-C (Closed Set): This dataset builds on CIFAR-100 by adding label noise. A label transition matrix $Q$ is utilized, with $Q_{ij} = Pr[\tilde y = j | y = i]$ denotes the likelihood of converting a clean label $y$ into a noisy label $\tilde y$. The noise rate $n_c$ varies between 0 and 1, and the noise type $\mathcal{T}$ can be either symmetric or asymmetric.

CIFAR80N-O (Open Set): In this dataset, the last 20 classes of CIFAR-100 are considered as OOD classes. Random noise based on $\mathcal{T}$ is then introduced to the labels of the remaining samples, with a noise ratio of $n_c$. This process results in an overall noise ratio of $n_{all} = 0.2 + 0.8n_c$.
\paragraph{WebFG-496}
WebFG-496 \citep{sun2021webly} is divided into three sub-datasets: Web-Airplan, Web-Bird, and Web-Car. It contains 53,339 web training images of 200 bird species (Web-bird), 100 aircraft types (Web-aircraft), and 196 car models (Web-car). Since the WebFG-496 dataset is obtained from the network, image labels may contain a certain level of noise. As a result, it is unnecessary to manually add label noise during the study.
\paragraph{Food101N}
Food101N \citep{bossard2014food} is a large-scale image dataset, obtained from the Internet. This dataset consists 310,009 images of food recipes classified in 101 classes. It is essential to point out that because images are collected from accessible web resources the images are not clean, and thus still contain some amount of noise labels. So that, this dataset does not require any manual label noise injections.
\paragraph{Experimental settings}

Following Jo-SRC \citep{yao2021jo}, we employ a 7-layer DNN network model for CIFAR100N-C and CIFAR80N-O. We use the Adam optimizer with a momentum of 0.9. The initial learning rate is set to 0.001, and the batch size is 128. The overall training consists of 300 epochs, with 10 epochs for pre-training and 290 for the main training phase. A linear decay of the learning rate to \(10^{-4}\) commences after 80 epochs. In the process of filtering clean samples, the relevant parameters \(\tau_s\) and \(\tau_h\) are set to 0.75 and 0.9. During the margin ranking function sample selection phase, the relevant parameters \(\tau_p\) is set to 0.9. The selection coefficients \(\lambda_1\) and \(\lambda_2\) are set to 0.05 eachare both set to 0.05.
For Web-Aircraft, Web-Bird, and Web-Car, we employ a pre-trained ResNet-50 model trained on ImageNet as our backbone. Network parameters are updated using the SGD optimizer with a momentum of 0.9, and the initial learning rate is set to 0.0005. Due to the larger image size in the Web-Car dataset and hardware limitations, the batch size for this experiment is limited to 16. The total training duration comprises 200 epochs, with 10 epochs designated for pre-training and 190 epochs for the main training phase. Learning rate decay is implemented through a cosine annealing schedule, commencing after the warmup period. All other parameters are maintained consistently with those used for the CIFAR datasets.
For Food101N, we follow the settings outlined in Jo-SRC, utilizing a pre-trained ResNet-50 for comparison. Parameters remain consistent with those employed on the CIFAR datasets.

\paragraph{Baselines}

We compare RSS-MGM (Algorithm \ref{alg:train}) with the following state-of-the-art algorithms. All methods are implemented in PyTorch with default parameters, and all experiments are conducted on an NVIDIA GeForce RTX 3090 GPU. To assess the performance of RSS-MGM on CIFAR100N-C and CIFAR80N-O, we compare RSS-MGM with the following state-of-the-art sample selection methods: Decoupling \citep{malach2017decoupling}, Co-teaching \citep{han2018co}, Co-teaching+ \citep{yu2019does}, JoCoR \citep{wei2020combating}, Jo-SRC \citep{yao2021jo}, and PNP \citep{sun2022pnp}.
To evaluate the effectiveness of RSS-MGM on Web-Aircraft, Web-Bird, and Web-Car, in addition to the above methods, we also compare it with the following methods: SELFIE \citep{song2019selfie}, PENCIL\citep{yi2019probabilistic}, AFM\citep{peng2020suppressing}, CRSSC\citep{sun2020crssc}, Self-adaptive\citep{huang2020self}, DivideMix\citep{li2020dividemix}, PLC\citep{zhang2021learning}, Peer-learning\citep{sun2021webly}.
For the evaluation on Food101N, we compare the following methods with ours: CleanNet \citep{lee2018cleannet}, DeepSelf \citep{han2019deep}, Jo-SRC \citep{yao2021jo} and PNP \citep{sun2022pnp}. Finally, direct training on noisy datasets serves as a straightforward baseline (referred to as Standard) for comparison.

\subsection{ The experiments on Synthetic Noisy Datasets}
\paragraph{Results on CIFAR100N-C.}
Although our method is intended to simulate real-world (open-set) noise situations, it is equally effective in closed-set situations. Table \ref{tab:tb1} shows the test accuracies for CIFAR100N-C. The results of existing approaches are sourced from Jo-SRC \citep{wei2020combating}, under same experimental settings as our approach. 

Under symmetric noise conditions, our method achieves remarkable average test accuracies of 64.6\%, 60.28\%, and 41.88\% at 20\%, 50\%, and 80\% noise rates, respectively. This demonstrates the adaptability of RSS-MGM to varied levels of symmetric noise. RSS-MGM outperforms the prior PNP approach (54.92\%) in a Symmetric-50\% noise environment, achieving an average test accuracy of 60.28\%. This outcome highlights the superiority of our method in dealing with common symmetric noise situations.

Even more remarkable is its performance in the presence of asymmetric noise. Our method achieves a high test accuracy of 64.06\% at Asymmetric-20\%. RSS-MGM performs well in the difficult Asymmetric-40\% noise environment, with an average test accuracy of 58.36\%. These findings demonstrate that RSS-MGM achieves strong performance not only under typical symmetric noise conditions but also in scenarios involving more challenging asymmetric noise. This robustness is critical for real-world applications that frequently confront complex and actual noise situations.

\paragraph{Results on CIFAR80N-O.}
CIFAR80N-O is specifically designed to simulate real-world (open-set) noise scenarios. In Table \ref{tab:tb2}, we provide a comparison between our method and state-of-the-art approaches. Similarly, the results of existing methods are derived from Jo-SRC \citep{wei2020combating}. Our methodology is assessed under identical experimental conditions.

Under Symmetric noise conditions, the method achieves average test accuracies of 67.32\%, 61.59\%, and 39.86\% for noise rate of 20\%, 50\%, and 80\%, respectively. These extraordinary results demonstrate our approach robustness in the face of varied degrees of symmetric noise, particularly in high-intensity noise environments where it maintains relatively high accuracy.

At 20\% Asymmetric noise, the approach achieves an average test accuracy of 61.23\%, significantly outperforming other methods and demonstrating its exceptional capabilities in mild non-symmetric noise conditions. Our method performs well in the complicated noise environment of Asymmetric-40\%, with an average test accuracy of 59.80\%. This result highlights the stability of RSS-MGM, which performs not only in symmetric noise but also in more challenging and real-world asymmetric noise issues.

\begin{table}[]
	\centering
	\caption{Average test accuracy (\%) on CIFAR100N-C over the last 10 epochs ("Sym" and "Asym" denote the symmetric and asymmetric label noise, respectively).}
	\label{tab:tb1}
	\begin{tabular}{rcccccc}
		\hline
		\multicolumn{1}{c}{\textbf{Methods}} & \textbf{Syms-20\%}       & \textbf{Syms-50\%}        & \textbf{Syms-80\%}        & \textbf{Asym-20\%}        & \textbf{Asym-40\%}        \\ \hline
		Standard                             & 35.14 $\pm$ 0.44         & 16.97 $\pm$ 0.40          & 4.41 $\pm$ 0.14           & 34.74 $\pm$ 0.53          & 27.29 $\pm$ 0.25          \\
		Decoupling\citep{malach2017decoupling}  & 33.10 $\pm$ 0.12         & 15.25 $\pm$ 0.20          & 3.89 $\pm$ 0.16           & 33.89 $\pm$ 0.23          & 26.11 $\pm$ 0.39          \\
		Co-teaching\citep{han2018co}            & 43.73 $\pm$ 0.16         & 34.96 $\pm$ 0.50          & 15.15 $\pm$ 0.46          & 35.82 $\pm$ 0.38          & 28.35 $\pm$ 0.25          \\
		Co-teaching+\citep{yu2019does}          & 49.27 $\pm$ 0.03         & 40.04 $\pm$ 0.70          & 13.44 $\pm$ 0.37          & 40.03 $\pm$ 0.64          & 33.62 $\pm$ 0.39          \\
		JoCoR\citep{wei2020combating}           & 53.01 $\pm$ 0.04         & 43.49 $\pm$ 0.46          & 15.49 $\pm$ 0.98          & 39.78 $\pm$ 0.38          & 32.70 $\pm$ 0.35          \\
		Jo-SRC\citep{yao2021jo}                 & 58.15 $\pm$ 0.14         & 51.26 $\pm$ 0.11          & 23.80 $\pm$ 0.05          & 63.96 $\pm$ 0.10          & 38.52 $\pm$ 0.20          \\
		PNP\citep{sun2022pnp}                   & 64.25 $\pm$ 0.12         & 53.74 $\pm$ 0.21          & 31.32 $\pm$ 0.19          & 63.65 $\pm$ 0.12          & \textbf{60.25 $\pm$ 0.21} \\ \hline
		\multicolumn{1}{c}{\textbf{RSS-MGM}} & \textbf{64.60 $\pm$ 0.08} & \textbf{60.28 $\pm$ 0.09} & \textbf{41.88 $\pm$ 0.12} & \textbf{64.06 $\pm$ 0.15} & 58.36 $\pm$ 0.13          \\ \hline
	\end{tabular}
\end{table}

\begin{table}[]
	\centering
	\caption{Average test accuracy (\%) on CIFAR80N-O over the last 10 epochs ("Sym" and "Asym" denote the symmetric and asymmetric label noise, respectively).}
	\label{tab:tb2}
	\begin{tabular}{rcccccc}
		\hline
		\multicolumn{1}{c}{\textbf{Methods}} & \textbf{Syms-20\%}        & \textbf{Syms-50\%}        & \textbf{Syms-80\%}        & \textbf{Asym-20\%}        & \textbf{Asym-40\%}        \\ \hline
		Standard                             & 29.37 $\pm$ 0.09          & 13.87 $\pm$ 0.08          & 4.20 $\pm$ 0.07           & 28.97 $\pm$ 0.04         & 22.25 $\pm$ 0.08          \\
		Decoupling\citep{malach2017decoupling}  & 43.49 $\pm$ 0.39          & 28.22 $\pm$ 0.19          & 10.01 $\pm$ 0.29          &  -                        & 33.74 $\pm$ 0.26          \\
		Co-teaching\citep{han2018co}            & 60.38 $\pm$ 0.22          & 52.42 $\pm$ 0.51          & 16.59 $\pm$ 0.27          &  -                        & 42.42 $\pm$ 0.30          \\
		Co-teaching+\citep{yu2019does}          & 53.97 $\pm$ 0.26          & 46.75 $\pm$ 0.14          & 12.29 $\pm$ 0.09          &  -                        & 43.01 $\pm$ 0.59          \\
		JoCoR\citep{wei2020combating}           & 59.99 $\pm$ 0.13          & 50.61 $\pm$ 0.12          & 12.85 $\pm$ 0.05          & 43.56 $\pm$ 0.25          & 39.37 $\pm$ 0.16          \\
		Jo-SRC\citep{yao2021jo}                 & 65.83 $\pm$ 0.13          & 58.51 $\pm$ 0.08          & 29.76 $\pm$ 0.09          & 66.96 $\pm$ 0.06          & 53.03 $\pm$ 0.25          \\
		PNP\citep{sun2022pnp}                   & 67.00 $\pm$ 0.18          & 54.92 $\pm$ 0.10          & 34.36 $\pm$ 0.18          & 65.51 $\pm$ 0.10          & \textbf{61.23 $\pm$ 0.17} \\ \hline
		\textbf{RSS-MGM}                     & \textbf{67.32 $\pm$ 0.15} & \textbf{61.59 $\pm$ 0.13} & \textbf{39.86 $\pm$ 0.08} & \textbf{67.02 $\pm$ 0.08} & 59.80 $\pm$ 0.16 \\ \hline
	\end{tabular}
\end{table}

\subsection{ The experiments on Real-world Noisy Datasets}
In addition to the evaluations shown above, we conduct experiments on real-world noisy datasets, such as the Web-Aircraft/Bird/Car and Food01N datasets, to validate the effectiveness of RSS-MGM.

\paragraph{Results on Web-Aircraft / Bird / Car.}
Web-Aircraft, Web-Bird, and Web-Car are three real-world web image datasets designed for fine-grained visual classification applications. These datasets differ from others in that over 25\% of the samples have unknown asymmetric noise labels, and no label validation information is provided. This feature makes these datasets more representative of real-world applications and provides an excellent experimental framework for investigating open-set noise label challenges. We validate our method using the same conditions as Jo-SRC's experimental methodology \citep{wei2020combating}. Table \ref{tab:tb3} shows an overview of test accuracies for Web-Aircraft, Web-Bird, and Web-Car with current methods.

\begin{table}[]
	\centering
	\caption{Comparison with state-of-the-art approaches in test accuracy (\%) on Web-Aircraft, Web-Bird, and Web-Car.}
	\label{tab:tb3}
	\begin{tabular}{rcccc}
		\hline
		\multicolumn{1}{c}{\textbf{Methods}} & \textbf{Web-Aircraft} & \textbf{Web-Bird} & \textbf{Web-Car} \\ \hline
		Standard                             & 60.80                 & 64.40             & 60.60            \\
		Decoupling\citep{malach2017decoupling} & 75.91                 & 71.61             & 79.41            \\
		Co-teaching\citep{han2018co}          & 79.54                 & 76.68             & 84.95            \\
		Co-teaching+\citep{yu2019does}        & 74.80                 & 70.12             & 76.77            \\
		SELFIE\citep{song2019selfie}          & 79.27                 & 77.20             & 82.90            \\
		PENCIL\citep{yi2019probabilistic}     & 78.82                 & 75.09             & 81.68            \\
		JoCoR\citep{wei2020combating}         & 80.11                 & 79.19             & 85.10            \\
		AFM\citep{peng2020suppressing}        & 81.04                 & 76.35             & 83.48            \\
		CRSSC\citep{sun2020crssc}             & 82.51                 & 81.31             & 87.68            \\
		Self-adaptive\citep{huang2020self}    & 77.92                 & 78.49             & 78.19            \\
		DivideMix\citep{li2020dividemix}      & 82.48                 & 74.40             & 84.27            \\
		Jo-SRC\citep{yao2021jo}               & 82.73                 & 81.22             & 88.13            \\
		PLC\citep{zhang2021learning}          & 79.24                 & 76.22             & 81.87            \\
		Peer-learning\citep{sun2021webly}     & 78.64                 & 75.37             & 82.48            \\
		PNP\citep{sun2022pnp}                 & 85.54                 & 81.93             & 90.11            \\ \hline
		\textbf{RSS-MGM}                     & \textbf{85.82}        & \textbf{82.34}    & \textbf{90.15}   \\ \hline
	\end{tabular}
\end{table}

\paragraph{Results on Food101N.}
As shown in Table \ref{tab:tb4}, the results of the study on the Food101N dataset show that RSS-MGM achieves higher test accuracy compared to existing state-of-the-art methods. These datasets have more complex label noise patterns than synthetic datasets, although at lower noise ratios. This complexity is a greater challenge. The results suggest that RSS-MGM can effectively reduce label noise in large-scale real-world applications, particularly in more complex and challenging settings.

\begin{table}[]
	\centering
	\caption{Comparison with state-of-the-art approaches in test accuracy (\%) on Food101N.}
	\label{tab:tb4}
	\begin{tabular}{rc}
		\hline
		\multicolumn{1}{c}{\textbf{Method}}  & \textbf{Test accuracy} \\ \hline
		Stardard                             & 84.51                  \\
		CleanNet\citep{lee2018cleannet}       & 83.47                  \\
		DeepSelf\citep{han2019deep}           & 85.11                  \\
		Jo-SRC\citep{yao2021jo}               & 86.66                  \\
		PNP\citep{sun2022pnp}                 & 87.50                  \\ \hline
		\multicolumn{1}{c}{\textbf{RSS-MGM}} & \textbf{88.73}         \\ \hline
	\end{tabular}
\end{table}

\subsection{ Ablation study}
To verify the effectiveness of every RSS-MGM module, we perform a set of ablation studies on the CIFAR80N-O dataset with a 20\% symmetric noise rate.
Firstly, the Robust Sample Selection module (RSS) and the Margin-Guided Module (MGM) are removed respectively, and then the two modules are removed at the same time. In addition, we conduct ablation study on Semi-Supervised Learning (SSL).

In its current condition, the RSS-MGM method achieves excellent results, with a test accuracy of 67.32\%. However, when the RSS is removed, the model's performance dropped slightly to 65.53\%. This indicates that RSS strongly selects clean samples, so decreasing interference of noisy data and improving model performance. Similarly, removing the MGM reduced test accuracy to 65.35\%, showing the importance of MGM in learning margin information and distinguishing between open set and closed set noise samples.
Further study shows that when RSS and MGM were removed at the same time, the model performance dropped to 62.47\%. This indicates the complementarity of the two modules in their combined role, which work together to increase the model's robustness and performance. Finally, to show the effect of semi-supervised learning, removing the SSL module resulted in a test accuracy of 66.49\%, which is 0.83\% lower than in the full situation. It emphasizes the positive influence of SSL in enhancing model performance while re-labeling the closed-set data and removing open-set data.
Table \ref{tab:tb5} shows the experimental results.

\begin{table}[]
	\centering
	\caption{Ablation study with sym 20 (\%) on CIFAR80N.}
	\label{tab:tb5}
	\begin{tabular}{lcc}
		\hline
		\textbf{Method}                                   & \textbf{Test accuracy(\%)} \\ \hline
		RSS-MGM                                        & 67.32                    \\
		RSS-MGM \emph{w/o} RSSM \& MGM                & 62.47                    \\
		RSS-MGM \emph{w/o} RSS                        & 65.53                    \\
		RSS-MGM \emph{w/o} MGS                        & 65.35                    \\
		RSS-MGM \emph{w/o} SSL                         & 66.49                    \\ \hline
	\end{tabular}
\end{table}
We design a set of ablation studies to show the necessity of two margin functions in the Margin-Guided Strategy Module. In these tests, we remove $\mathcal{M}_v$ and $\mathcal{M}_p$, respectively, to get insight into their impact to the performance of RSS-MGM. Table \ref{tab:tb6} displays the results of our ablation studies conducted on the CIFAR80N-O dataset with a 20\% symmetric noise rate.

\begin{table}[]
	\centering
	\caption{Ablation study with sym 20 (\%) on CIFAR80N.}
	\label{tab:tb6}
	\begin{tabular}{cccc}
		\hline
		\textbf{Method}                                                      & \textbf{Test accuracy(\%)} \\ \hline
		RSS-MGM                                                            &67.32                    \\
		RSS-MGM \emph{w/o} $\mathcal{M}_v$                                 &64.75                    \\
		RSS-MGM \emph{w/o} $\mathcal{M}_p$                                 &65.38                    \\ \hline
	\end{tabular}
\end{table}

\section{Conclusion}\label{sec5}
In this research, we focus on tackling the difficulties of learning from real-world datasets that include both closed-set and open-set label noise. To deal with this problem, RSS-MGM is proposed in this paper, which employs a multi-level sample selection method to divide the training dataset that goes beyond simply deleting noisy labels. Instead, it reduces the interference from noisy labels while keeping data features.
This method is developed with the complexity of real-world datasets, which is critical in practical applications.
In the experimental section, we compare the performance of RSS-MGM on synthetic and real-world datasets, which outperforms many state-of-the-art methods.
In conclusion, the RSS-MGM method provides satisfactory results when dealing with the open set noise label problem.

\section{Conflict of Interest}
The authors declare that they have no conflict of interest.

\section*{Acknowledgments}
This research is supported by the National Key R \& D Program of China(No. 2021YFA1003004),the Open Fund for Key Laboratory of Internet of Aircrafts (No. MHFLW202305), the Tianjin Education Commission Research Program Project (2023KJ239) and the Large-scale Numerical Simulation Computing Sharing Platform of Shanghai University.

\section*{Data Availability}

The CIFAR-100 dataset is availiable at https://www.cs.toronto.edu/~kriz/cifar.html.
The WebFG-469 dataset is availiable at https://github.com/NUST-Machine-Intelligence-Laboratory/weblyFG-dataset.
The Food101N dataset is availiable at https://kuanghuei.github.io/Food-101N/.

%%       \citet{<label>} ==> Jones et al. [21]
%%       \citep{<label>} ==> [21]
%%

\appendix


\bibliographystyle{elsarticle-num-names} 
\bibliography{cas-refs}

\begin{thebibliography}{77}
\expandafter\ifx\csname natexlab\endcsname\relax\def\natexlab#1{#1}\fi
\providecommand{\url}[1]{\texttt{#1}}
\providecommand{\href}[2]{#2}
\providecommand{\path}[1]{#1}
\providecommand{\DOIprefix}{doi:}
\providecommand{\ArXivprefix}{arXiv:}
\providecommand{\URLprefix}{URL: }
\providecommand{\Pubmedprefix}{pmid:}
\providecommand{\doi}[1]{\href{http://dx.doi.org/#1}{\path{#1}}}
\providecommand{\Pubmed}[1]{\href{pmid:#1}{\path{#1}}}
\providecommand{\bibinfo}[2]{#2}
\ifx\xfnm\relax \def\xfnm[#1]{\unskip,\space#1}\fi
%Type = Article
\bibitem[{Krizhevsky et~al.(2012)Krizhevsky, Sutskever, and Hinton}]{2012ImageNet}
\bibinfo{author}{A.~Krizhevsky}, \bibinfo{author}{I.~Sutskever}, \bibinfo{author}{G.~Hinton},
\newblock \bibinfo{title}{Imagenet classification with deep convolutional neural networks},
\newblock \bibinfo{journal}{Advances in neural information processing systems} \bibinfo{volume}{25} (\bibinfo{year}{2012}).
%Type = Article
\bibitem[{Ren et~al.(2015)Ren, He, Girshick, and Sun}]{ren2015faster}
\bibinfo{author}{S.~Ren}, \bibinfo{author}{K.~He}, \bibinfo{author}{R.~Girshick}, \bibinfo{author}{J.~Sun},
\newblock \bibinfo{title}{Faster r-cnn: Towards real-time object detection with region proposal networks},
\newblock \bibinfo{journal}{Advances in neural information processing systems} \bibinfo{volume}{28} (\bibinfo{year}{2015}).
%Type = Inproceedings
\bibitem[{Redmon and Farhadi(2017)}]{redmon2017yolo9000}
\bibinfo{author}{J.~Redmon}, \bibinfo{author}{A.~Farhadi},
\newblock \bibinfo{title}{Yolo9000: better, faster, stronger},
\newblock in: \bibinfo{booktitle}{Proceedings of the IEEE conference on computer vision and pattern recognition}, \bibinfo{year}{2017}, pp. \bibinfo{pages}{7263--7271}.
%Type = Inproceedings
\bibitem[{Xie et~al.(2020)Xie, Liu, Zhu, Zhao, Zhang, Yao, Qin, and Shao}]{xie2020region}
\bibinfo{author}{G.-S. Xie}, \bibinfo{author}{L.~Liu}, \bibinfo{author}{F.~Zhu}, \bibinfo{author}{F.~Zhao}, \bibinfo{author}{Z.~Zhang}, \bibinfo{author}{Y.~Yao}, \bibinfo{author}{J.~Qin}, \bibinfo{author}{L.~Shao},
\newblock \bibinfo{title}{Region graph embedding network for zero-shot learning},
\newblock in: \bibinfo{booktitle}{Computer Vision--ECCV 2020: 16th European Conference, Glasgow, UK, August 23--28, 2020, Proceedings, Part IV 16}, \bibinfo{organization}{Springer}, \bibinfo{year}{2020}, pp. \bibinfo{pages}{562--580}.
%Type = Inproceedings
\bibitem[{Luo et~al.(2019)Luo, Lin, Liu, Liu, Tang, and Yao}]{luo2019segeqa}
\bibinfo{author}{H.~Luo}, \bibinfo{author}{G.~Lin}, \bibinfo{author}{Z.~Liu}, \bibinfo{author}{F.~Liu}, \bibinfo{author}{Z.~Tang}, \bibinfo{author}{Y.~Yao},
\newblock \bibinfo{title}{Segeqa: Video segmentation based visual attention for embodied question answering},
\newblock in: \bibinfo{booktitle}{2019 IEEE/CVF International Conference on Computer Vision (ICCV)}, \bibinfo{organization}{IEEE}, \bibinfo{year}{2019}, pp. \bibinfo{pages}{9666--9675}.
%Type = Inproceedings
\bibitem[{Deng et~al.(2009)Deng, Dong, Socher, Li, Li, and Fei-Fei}]{deng2009imagenet}
\bibinfo{author}{J.~Deng}, \bibinfo{author}{W.~Dong}, \bibinfo{author}{R.~Socher}, \bibinfo{author}{L.-J. Li}, \bibinfo{author}{K.~Li}, \bibinfo{author}{L.~Fei-Fei},
\newblock \bibinfo{title}{Imagenet: A large-scale hierarchical image database},
\newblock in: \bibinfo{booktitle}{2009 IEEE conference on computer vision and pattern recognition}, \bibinfo{organization}{Ieee}, \bibinfo{year}{2009}, pp. \bibinfo{pages}{248--255}.
%Type = Inproceedings
\bibitem[{Lin et~al.(2014)Lin, Maire, Belongie, Hays, Perona, Ramanan, Doll{\'a}r, and Zitnick}]{lin2014microsoft}
\bibinfo{author}{T.-Y. Lin}, \bibinfo{author}{M.~Maire}, \bibinfo{author}{S.~Belongie}, \bibinfo{author}{J.~Hays}, \bibinfo{author}{P.~Perona}, \bibinfo{author}{D.~Ramanan}, \bibinfo{author}{P.~Doll{\'a}r}, \bibinfo{author}{C.~L. Zitnick},
\newblock \bibinfo{title}{Microsoft coco: Common objects in context},
\newblock in: \bibinfo{booktitle}{Computer Vision--ECCV 2014: 13th European Conference, Zurich, Switzerland, September 6-12, 2014, Proceedings, Part V 13}, \bibinfo{organization}{Springer}, \bibinfo{year}{2014}, pp. \bibinfo{pages}{740--755}.
%Type = Article
\bibitem[{Li et~al.(2020)Li, Socher, and Hoi}]{li2020dividemix}
\bibinfo{author}{J.~Li}, \bibinfo{author}{R.~Socher}, \bibinfo{author}{S.~C. Hoi},
\newblock \bibinfo{title}{Dividemix: Learning with noisy labels as semi-supervised learning},
\newblock \bibinfo{journal}{arXiv preprint arXiv:2002.07394}  (\bibinfo{year}{2020}).
%Type = Article
\bibitem[{Wang et~al.(2022)Wang, Xiao, Dong, Feng, and Zhao}]{wang2022promix}
\bibinfo{author}{H.~Wang}, \bibinfo{author}{R.~Xiao}, \bibinfo{author}{Y.~Dong}, \bibinfo{author}{L.~Feng}, \bibinfo{author}{J.~Zhao},
\newblock \bibinfo{title}{Promix: combating label noise via maximizing clean sample utility},
\newblock \bibinfo{journal}{arXiv preprint arXiv:2207.10276}  (\bibinfo{year}{2022}).
%Type = Article
\bibitem[{Wu et~al.(2020)Wu, Xia, Liu, Han, Gong, Wang, Liu, and Niu}]{wu2020class2simi}
\bibinfo{author}{S.~Wu}, \bibinfo{author}{X.~Xia}, \bibinfo{author}{T.~Liu}, \bibinfo{author}{B.~Han}, \bibinfo{author}{M.~Gong}, \bibinfo{author}{N.~Wang}, \bibinfo{author}{H.~Liu}, \bibinfo{author}{G.~Niu},
\newblock \bibinfo{title}{Class2simi: A new perspective on learning with label noise}  (\bibinfo{year}{2020}).
%Type = Inproceedings
\bibitem[{Sehwag et~al.(2019)Sehwag, Bhagoji, Song, Sitawarin, Cullina, Chiang, and Mittal}]{sehwag2019analyzing}
\bibinfo{author}{V.~Sehwag}, \bibinfo{author}{A.~N. Bhagoji}, \bibinfo{author}{L.~Song}, \bibinfo{author}{C.~Sitawarin}, \bibinfo{author}{D.~Cullina}, \bibinfo{author}{M.~Chiang}, \bibinfo{author}{P.~Mittal},
\newblock \bibinfo{title}{Analyzing the robustness of open-world machine learning},
\newblock in: \bibinfo{booktitle}{Proceedings of the 12th ACM Workshop on Artificial Intelligence and Security}, \bibinfo{year}{2019}, pp. \bibinfo{pages}{105--116}.
%Type = Inproceedings
\bibitem[{Sun et~al.(2021)Sun, Yao, Wei, Zhang, Shen, Wu, Zhang, and Shen}]{sun2021webly}
\bibinfo{author}{Z.~Sun}, \bibinfo{author}{Y.~Yao}, \bibinfo{author}{X.-S. Wei}, \bibinfo{author}{Y.~Zhang}, \bibinfo{author}{F.~Shen}, \bibinfo{author}{J.~Wu}, \bibinfo{author}{J.~Zhang}, \bibinfo{author}{H.~T. Shen},
\newblock \bibinfo{title}{Webly supervised fine-grained recognition: Benchmark datasets and an approach},
\newblock in: \bibinfo{booktitle}{Proceedings of the IEEE/CVF international conference on computer vision}, \bibinfo{year}{2021}, pp. \bibinfo{pages}{10602--10611}.
%Type = Article
\bibitem[{Yang et~al.(2018)Yang, Sun, Lai, Zheng, and Cheng}]{yang2018recognition}
\bibinfo{author}{J.~Yang}, \bibinfo{author}{X.~Sun}, \bibinfo{author}{Y.-K. Lai}, \bibinfo{author}{L.~Zheng}, \bibinfo{author}{M.-M. Cheng},
\newblock \bibinfo{title}{Recognition from web data: A progressive filtering approach},
\newblock \bibinfo{journal}{IEEE Transactions on Image Processing} \bibinfo{volume}{27} (\bibinfo{year}{2018}) \bibinfo{pages}{5303--5315}.
%Type = Inproceedings
\bibitem[{Yao et~al.(2020)Yao, Hua, Gao, Sun, Li, and Zhang}]{yao2020bridging}
\bibinfo{author}{Y.~Yao}, \bibinfo{author}{X.~Hua}, \bibinfo{author}{G.~Gao}, \bibinfo{author}{Z.~Sun}, \bibinfo{author}{Z.~Li}, \bibinfo{author}{J.~Zhang},
\newblock \bibinfo{title}{Bridging the web data and fine-grained visual recognition via alleviating label noise and domain mismatch},
\newblock in: \bibinfo{booktitle}{Proceedings of the 28th ACM International Conference on Multimedia}, \bibinfo{year}{2020}, pp. \bibinfo{pages}{1735--1744}.
%Type = Inproceedings
\bibitem[{Tanaka et~al.(2018)Tanaka, Ikami, Yamasaki, and Aizawa}]{tanaka2018joint}
\bibinfo{author}{D.~Tanaka}, \bibinfo{author}{D.~Ikami}, \bibinfo{author}{T.~Yamasaki}, \bibinfo{author}{K.~Aizawa},
\newblock \bibinfo{title}{Joint optimization framework for learning with noisy labels},
\newblock in: \bibinfo{booktitle}{Proceedings of the IEEE conference on computer vision and pattern recognition}, \bibinfo{year}{2018}, pp. \bibinfo{pages}{5552--5560}.
%Type = Article
\bibitem[{Yao et~al.(2018)Yao, Shen, Zhang, Liu, Tang, and Shao}]{yao2018extracting}
\bibinfo{author}{Y.~Yao}, \bibinfo{author}{F.~Shen}, \bibinfo{author}{J.~Zhang}, \bibinfo{author}{L.~Liu}, \bibinfo{author}{Z.~Tang}, \bibinfo{author}{L.~Shao},
\newblock \bibinfo{title}{Extracting multiple visual senses for web learning},
\newblock \bibinfo{journal}{IEEE Transactions on Multimedia} \bibinfo{volume}{21} (\bibinfo{year}{2018}) \bibinfo{pages}{184--196}.
%Type = Inproceedings
\bibitem[{Zhang et~al.(2019)Zhang, Wang, and Qiao}]{zhang2019metacleaner}
\bibinfo{author}{W.~Zhang}, \bibinfo{author}{Y.~Wang}, \bibinfo{author}{Y.~Qiao},
\newblock \bibinfo{title}{Metacleaner: Learning to hallucinate clean representations for noisy-labeled visual recognition},
\newblock in: \bibinfo{booktitle}{Proceedings of the IEEE/CVF Conference on Computer Vision and Pattern Recognition}, \bibinfo{year}{2019}, pp. \bibinfo{pages}{7373--7382}.
%Type = Article
\bibitem[{Liu et~al.(2021)Liu, Zhang, Yao, Wei, Shen, Tang, and Zhang}]{liu2021exploiting}
\bibinfo{author}{H.~Liu}, \bibinfo{author}{C.~Zhang}, \bibinfo{author}{Y.~Yao}, \bibinfo{author}{X.-S. Wei}, \bibinfo{author}{F.~Shen}, \bibinfo{author}{Z.~Tang}, \bibinfo{author}{J.~Zhang},
\newblock \bibinfo{title}{Exploiting web images for fine-grained visual recognition by eliminating open-set noise and utilizing hard examples},
\newblock \bibinfo{journal}{IEEE Transactions on Multimedia} \bibinfo{volume}{24} (\bibinfo{year}{2021}) \bibinfo{pages}{546--557}.
%Type = Article
\bibitem[{Sun et~al.(2021)Sun, Liu, Wang, Zhou, Wu, and Tang}]{sun2021co}
\bibinfo{author}{Z.~Sun}, \bibinfo{author}{H.~Liu}, \bibinfo{author}{Q.~Wang}, \bibinfo{author}{T.~Zhou}, \bibinfo{author}{Q.~Wu}, \bibinfo{author}{Z.~Tang},
\newblock \bibinfo{title}{Co-ldl: A co-training-based label distribution learning method for tackling label noise},
\newblock \bibinfo{journal}{IEEE Transactions on Multimedia} \bibinfo{volume}{24} (\bibinfo{year}{2021}) \bibinfo{pages}{1093--1104}.
%Type = Article
\bibitem[{Li et~al.(2017)Li, Wang, Li, Agustsson, and Van~Gool}]{li2017webvision}
\bibinfo{author}{W.~Li}, \bibinfo{author}{L.~Wang}, \bibinfo{author}{W.~Li}, \bibinfo{author}{E.~Agustsson}, \bibinfo{author}{L.~Van~Gool},
\newblock \bibinfo{title}{Webvision database: Visual learning and understanding from web data},
\newblock \bibinfo{journal}{arXiv preprint arXiv:1708.02862}  (\bibinfo{year}{2017}).
%Type = Inproceedings
\bibitem[{Xiao et~al.(2015)Xiao, Xia, Yang, Huang, and Wang}]{xiao2015learning}
\bibinfo{author}{T.~Xiao}, \bibinfo{author}{T.~Xia}, \bibinfo{author}{Y.~Yang}, \bibinfo{author}{C.~Huang}, \bibinfo{author}{X.~Wang},
\newblock \bibinfo{title}{Learning from massive noisy labeled data for image classification},
\newblock in: \bibinfo{booktitle}{Proceedings of the IEEE conference on computer vision and pattern recognition}, \bibinfo{year}{2015}, pp. \bibinfo{pages}{2691--2699}.
%Type = Inproceedings
\bibitem[{Sun et~al.(2020)Sun, Hua, Yao, Wei, Hu, and Zhang}]{sun2020crssc}
\bibinfo{author}{Z.~Sun}, \bibinfo{author}{X.-S. Hua}, \bibinfo{author}{Y.~Yao}, \bibinfo{author}{X.-S. Wei}, \bibinfo{author}{G.~Hu}, \bibinfo{author}{J.~Zhang},
\newblock \bibinfo{title}{Crssc: salvage reusable samples from noisy data for robust learning},
\newblock in: \bibinfo{booktitle}{Proceedings of the 28th ACM International Conference on Multimedia}, \bibinfo{year}{2020}, pp. \bibinfo{pages}{92--101}.
%Type = Article
\bibitem[{Yao et~al.(2019)Yao, Zhang, Shen, Liu, Zhu, Zhang, and Shen}]{yao2019towards}
\bibinfo{author}{Y.~Yao}, \bibinfo{author}{J.~Zhang}, \bibinfo{author}{F.~Shen}, \bibinfo{author}{L.~Liu}, \bibinfo{author}{F.~Zhu}, \bibinfo{author}{D.~Zhang}, \bibinfo{author}{H.~T. Shen},
\newblock \bibinfo{title}{Towards automatic construction of diverse, high-quality image datasets},
\newblock \bibinfo{journal}{IEEE Transactions on Knowledge and Data Engineering} \bibinfo{volume}{32} (\bibinfo{year}{2019}) \bibinfo{pages}{1199--1211}.
%Type = Article
\bibitem[{Nettleton et~al.(2010)Nettleton, Orriols-Puig, and Fornells}]{nettleton2010study}
\bibinfo{author}{D.~F. Nettleton}, \bibinfo{author}{A.~Orriols-Puig}, \bibinfo{author}{A.~Fornells},
\newblock \bibinfo{title}{A study of the effect of different types of noise on the precision of supervised learning techniques},
\newblock \bibinfo{journal}{Artificial intelligence review} \bibinfo{volume}{33} (\bibinfo{year}{2010}) \bibinfo{pages}{275--306}.
%Type = Article
\bibitem[{Zhang et~al.(2021)Zhang, Bengio, Hardt, Recht, and Vinyals}]{zhang2021understanding}
\bibinfo{author}{C.~Zhang}, \bibinfo{author}{S.~Bengio}, \bibinfo{author}{M.~Hardt}, \bibinfo{author}{B.~Recht}, \bibinfo{author}{O.~Vinyals},
\newblock \bibinfo{title}{Understanding deep learning (still) requires rethinking generalization},
\newblock \bibinfo{journal}{Communications of the ACM} \bibinfo{volume}{64} (\bibinfo{year}{2021}) \bibinfo{pages}{107--115}.
%Type = Inproceedings
\bibitem[{Arpit et~al.(2017)Arpit, Jastrz{\k{e}}bski, Ballas, Krueger, Bengio, Kanwal, Maharaj, Fischer, Courville, Bengio et~al.}]{arpit2017closer}
\bibinfo{author}{D.~Arpit}, \bibinfo{author}{S.~Jastrz{\k{e}}bski}, \bibinfo{author}{N.~Ballas}, \bibinfo{author}{D.~Krueger}, \bibinfo{author}{E.~Bengio}, \bibinfo{author}{M.~S. Kanwal}, \bibinfo{author}{T.~Maharaj}, \bibinfo{author}{A.~Fischer}, \bibinfo{author}{A.~Courville}, \bibinfo{author}{Y.~Bengio}, et~al.,
\newblock \bibinfo{title}{A closer look at memorization in deep networks},
\newblock in: \bibinfo{booktitle}{International conference on machine learning}, \bibinfo{organization}{PMLR}, \bibinfo{year}{2017}, pp. \bibinfo{pages}{233--242}.
%Type = Inproceedings
\bibitem[{Wang et~al.(2018)Wang, Liu, Ma, Bailey, Zha, Song, and Xia}]{wang2018iterative}
\bibinfo{author}{Y.~Wang}, \bibinfo{author}{W.~Liu}, \bibinfo{author}{X.~Ma}, \bibinfo{author}{J.~Bailey}, \bibinfo{author}{H.~Zha}, \bibinfo{author}{L.~Song}, \bibinfo{author}{S.-T. Xia},
\newblock \bibinfo{title}{Iterative learning with open-set noisy labels},
\newblock in: \bibinfo{booktitle}{Proceedings of the IEEE conference on computer vision and pattern recognition}, \bibinfo{year}{2018}, pp. \bibinfo{pages}{8688--8696}.
%Type = Inproceedings
\bibitem[{Lin et~al.(2017)Lin, Goyal, Girshick, He, and Doll{\'a}r}]{lin2017focal}
\bibinfo{author}{T.-Y. Lin}, \bibinfo{author}{P.~Goyal}, \bibinfo{author}{R.~Girshick}, \bibinfo{author}{K.~He}, \bibinfo{author}{P.~Doll{\'a}r},
\newblock \bibinfo{title}{Focal loss for dense object detection},
\newblock in: \bibinfo{booktitle}{Proceedings of the IEEE international conference on computer vision}, \bibinfo{year}{2017}, pp. \bibinfo{pages}{2980--2988}.
%Type = Inproceedings
\bibitem[{Ghosh et~al.(2017)Ghosh, Kumar, and Sastry}]{ghosh2017robust}
\bibinfo{author}{A.~Ghosh}, \bibinfo{author}{H.~Kumar}, \bibinfo{author}{P.~S. Sastry},
\newblock \bibinfo{title}{Robust loss functions under label noise for deep neural networks},
\newblock in: \bibinfo{booktitle}{Proceedings of the AAAI conference on artificial intelligence}, volume~\bibinfo{volume}{31}, \bibinfo{year}{2017}.
%Type = Article
\bibitem[{Zhang and Sabuncu(2018)}]{zhang2018generalized}
\bibinfo{author}{Z.~Zhang}, \bibinfo{author}{M.~Sabuncu},
\newblock \bibinfo{title}{Generalized cross entropy loss for training deep neural networks with noisy labels},
\newblock \bibinfo{journal}{Advances in neural information processing systems} \bibinfo{volume}{31} (\bibinfo{year}{2018}).
%Type = Article
\bibitem[{Wu et~al.(2018)Wu, He, Sun, and Tan}]{wu2018light}
\bibinfo{author}{X.~Wu}, \bibinfo{author}{R.~He}, \bibinfo{author}{Z.~Sun}, \bibinfo{author}{T.~Tan},
\newblock \bibinfo{title}{A light cnn for deep face representation with noisy labels},
\newblock \bibinfo{journal}{IEEE Transactions on Information Forensics and Security} \bibinfo{volume}{13} (\bibinfo{year}{2018}) \bibinfo{pages}{2884--2896}.
%Type = Article
\bibitem[{Reed et~al.(2014)Reed, Lee, Anguelov, Szegedy, Erhan, and Rabinovich}]{reed2014training}
\bibinfo{author}{S.~Reed}, \bibinfo{author}{H.~Lee}, \bibinfo{author}{D.~Anguelov}, \bibinfo{author}{C.~Szegedy}, \bibinfo{author}{D.~Erhan}, \bibinfo{author}{A.~Rabinovich},
\newblock \bibinfo{title}{Training deep neural networks on noisy labels with bootstrapping},
\newblock \bibinfo{journal}{arXiv preprint arXiv:1412.6596}  (\bibinfo{year}{2014}).
%Type = Inproceedings
\bibitem[{Arazo et~al.(2019)Arazo, Ortego, Albert, O’Connor, and McGuinness}]{arazo2019unsupervised}
\bibinfo{author}{E.~Arazo}, \bibinfo{author}{D.~Ortego}, \bibinfo{author}{P.~Albert}, \bibinfo{author}{N.~O’Connor}, \bibinfo{author}{K.~McGuinness},
\newblock \bibinfo{title}{Unsupervised label noise modeling and loss correction},
\newblock in: \bibinfo{booktitle}{International conference on machine learning}, \bibinfo{organization}{PMLR}, \bibinfo{year}{2019}, pp. \bibinfo{pages}{312--321}.
%Type = Inproceedings
\bibitem[{Zheng et~al.(2020)Zheng, Wu, Goswami, Goswami, Metaxas, and Chen}]{zheng2020error}
\bibinfo{author}{S.~Zheng}, \bibinfo{author}{P.~Wu}, \bibinfo{author}{A.~Goswami}, \bibinfo{author}{M.~Goswami}, \bibinfo{author}{D.~Metaxas}, \bibinfo{author}{C.~Chen},
\newblock \bibinfo{title}{Error-bounded correction of noisy labels},
\newblock in: \bibinfo{booktitle}{International Conference on Machine Learning}, \bibinfo{organization}{PMLR}, \bibinfo{year}{2020}, pp. \bibinfo{pages}{11447--11457}.
%Type = Article
\bibitem[{Zhang et~al.(2017)Zhang, Cisse, Dauphin, and Lopez-Paz}]{zhang2017mixup}
\bibinfo{author}{H.~Zhang}, \bibinfo{author}{M.~Cisse}, \bibinfo{author}{Y.~N. Dauphin}, \bibinfo{author}{D.~Lopez-Paz},
\newblock \bibinfo{title}{mixup: Beyond empirical risk minimization},
\newblock \bibinfo{journal}{arXiv preprint arXiv:1710.09412}  (\bibinfo{year}{2017}).
%Type = Article
\bibitem[{Han et~al.(2018)Han, Yao, Yu, Niu, Xu, Hu, Tsang, and Sugiyama}]{han2018co}
\bibinfo{author}{B.~Han}, \bibinfo{author}{Q.~Yao}, \bibinfo{author}{X.~Yu}, \bibinfo{author}{G.~Niu}, \bibinfo{author}{M.~Xu}, \bibinfo{author}{W.~Hu}, \bibinfo{author}{I.~Tsang}, \bibinfo{author}{M.~Sugiyama},
\newblock \bibinfo{title}{Co-teaching: Robust training of deep neural networks with extremely noisy labels},
\newblock \bibinfo{journal}{Advances in neural information processing systems} \bibinfo{volume}{31} (\bibinfo{year}{2018}).
%Type = Inproceedings
\bibitem[{Lee et~al.(2018)Lee, He, Zhang, and Yang}]{lee2018cleannet}
\bibinfo{author}{K.-H. Lee}, \bibinfo{author}{X.~He}, \bibinfo{author}{L.~Zhang}, \bibinfo{author}{L.~Yang},
\newblock \bibinfo{title}{Cleannet: Transfer learning for scalable image classifier training with label noise},
\newblock in: \bibinfo{booktitle}{Proceedings of the IEEE conference on computer vision and pattern recognition}, \bibinfo{year}{2018}, pp. \bibinfo{pages}{5447--5456}.
%Type = Article
\bibitem[{Xia et~al.(2021)Xia, Liu, Han, Gong, Yu, Niu, and Sugiyama}]{xia2021instance}
\bibinfo{author}{X.~Xia}, \bibinfo{author}{T.~Liu}, \bibinfo{author}{B.~Han}, \bibinfo{author}{M.~Gong}, \bibinfo{author}{J.~Yu}, \bibinfo{author}{G.~Niu}, \bibinfo{author}{M.~Sugiyama},
\newblock \bibinfo{title}{Instance correction for learning with open-set noisy labels},
\newblock \bibinfo{journal}{arXiv preprint arXiv:2106.00455}  (\bibinfo{year}{2021}).
%Type = Inproceedings
\bibitem[{Sachdeva et~al.(2021)Sachdeva, Cordeiro, Belagiannis, Reid, and Carneiro}]{sachdeva2021evidentialmix}
\bibinfo{author}{R.~Sachdeva}, \bibinfo{author}{F.~R. Cordeiro}, \bibinfo{author}{V.~Belagiannis}, \bibinfo{author}{I.~Reid}, \bibinfo{author}{G.~Carneiro},
\newblock \bibinfo{title}{Evidentialmix: Learning with combined open-set and closed-set noisy labels},
\newblock in: \bibinfo{booktitle}{Proceedings of the IEEE/CVF Winter Conference on Applications of Computer Vision}, \bibinfo{year}{2021}, pp. \bibinfo{pages}{3607--3615}.
%Type = Article
\bibitem[{Li et~al.(2020)Li, Xiong, and Hoi}]{li2020mopro}
\bibinfo{author}{J.~Li}, \bibinfo{author}{C.~Xiong}, \bibinfo{author}{S.~C. Hoi},
\newblock \bibinfo{title}{Mopro: Webly supervised learning with momentum prototypes},
\newblock \bibinfo{journal}{arXiv preprint arXiv:2009.07995}  (\bibinfo{year}{2020}).
%Type = Inproceedings
\bibitem[{Yao et~al.(2021)Yao, Sun, Zhang, Shen, Wu, Zhang, and Tang}]{yao2021jo}
\bibinfo{author}{Y.~Yao}, \bibinfo{author}{Z.~Sun}, \bibinfo{author}{C.~Zhang}, \bibinfo{author}{F.~Shen}, \bibinfo{author}{Q.~Wu}, \bibinfo{author}{J.~Zhang}, \bibinfo{author}{Z.~Tang},
\newblock \bibinfo{title}{Jo-src: A contrastive approach for combating noisy labels},
\newblock in: \bibinfo{booktitle}{Proceedings of the IEEE/CVF conference on computer vision and pattern recognition}, \bibinfo{year}{2021}, pp. \bibinfo{pages}{5192--5201}.
%Type = Inproceedings
\bibitem[{Sun et~al.(2022)Sun, Shen, Huang, Wang, Shu, Yao, and Tang}]{sun2022pnp}
\bibinfo{author}{Z.~Sun}, \bibinfo{author}{F.~Shen}, \bibinfo{author}{D.~Huang}, \bibinfo{author}{Q.~Wang}, \bibinfo{author}{X.~Shu}, \bibinfo{author}{Y.~Yao}, \bibinfo{author}{J.~Tang},
\newblock \bibinfo{title}{Pnp: Robust learning from noisy labels by probabilistic noise prediction},
\newblock in: \bibinfo{booktitle}{Proceedings of the IEEE/CVF Conference on Computer Vision and Pattern Recognition}, \bibinfo{year}{2022}, pp. \bibinfo{pages}{5311--5320}.
%Type = Article
\bibitem[{Cai et~al.(2023)Cai, Liu, Huang, Yao, and Tang}]{cai2023co}
\bibinfo{author}{Z.~Cai}, \bibinfo{author}{H.~Liu}, \bibinfo{author}{D.~Huang}, \bibinfo{author}{Y.~Yao}, \bibinfo{author}{Z.~Tang},
\newblock \bibinfo{title}{Co-mining: Mining informative samples with noisy labels},
\newblock \bibinfo{journal}{Signal Processing} \bibinfo{volume}{209} (\bibinfo{year}{2023}) \bibinfo{pages}{109003}.
%Type = Inproceedings
\bibitem[{Albert et~al.(2023)Albert, Arazo, Krishna, O’Connor, and McGuinness}]{albert2023your}
\bibinfo{author}{P.~Albert}, \bibinfo{author}{E.~Arazo}, \bibinfo{author}{T.~Krishna}, \bibinfo{author}{N.~E. O’Connor}, \bibinfo{author}{K.~McGuinness},
\newblock \bibinfo{title}{Is your noise correction noisy? pls: Robustness to label noise with two stage detection},
\newblock in: \bibinfo{booktitle}{Proceedings of the IEEE/CVF Winter Conference on Applications of Computer Vision}, \bibinfo{year}{2023}, pp. \bibinfo{pages}{118--127}.
%Type = Article
\bibitem[{Xu et~al.(2023)Xu, Lin, Cai, and Wang}]{xu2023label}
\bibinfo{author}{C.~Xu}, \bibinfo{author}{R.~Lin}, \bibinfo{author}{J.~Cai}, \bibinfo{author}{S.~Wang},
\newblock \bibinfo{title}{Label correction using contrastive prototypical classifier for noisy label learning},
\newblock \bibinfo{journal}{Information Sciences} \bibinfo{volume}{649} (\bibinfo{year}{2023}) \bibinfo{pages}{119647}.
%Type = Inproceedings
\bibitem[{Yu et~al.(2019)Yu, Han, Yao, Niu, Tsang, and Sugiyama}]{yu2019does}
\bibinfo{author}{X.~Yu}, \bibinfo{author}{B.~Han}, \bibinfo{author}{J.~Yao}, \bibinfo{author}{G.~Niu}, \bibinfo{author}{I.~Tsang}, \bibinfo{author}{M.~Sugiyama},
\newblock \bibinfo{title}{How does disagreement help generalization against label corruption?},
\newblock in: \bibinfo{booktitle}{International Conference on Machine Learning}, \bibinfo{organization}{PMLR}, \bibinfo{year}{2019}, pp. \bibinfo{pages}{7164--7173}.
%Type = Article
\bibitem[{Pleiss et~al.(2020)Pleiss, Zhang, Elenberg, and Weinberger}]{pleiss2020identifying}
\bibinfo{author}{G.~Pleiss}, \bibinfo{author}{T.~Zhang}, \bibinfo{author}{E.~Elenberg}, \bibinfo{author}{K.~Q. Weinberger},
\newblock \bibinfo{title}{Identifying mislabeled data using the area under the margin ranking},
\newblock \bibinfo{journal}{Advances in Neural Information Processing Systems} \bibinfo{volume}{33} (\bibinfo{year}{2020}) \bibinfo{pages}{17044--17056}.
%Type = Article
\bibitem[{Cheng et~al.(2020)Cheng, Zhu, Li, Gong, Sun, and Liu}]{cheng2020learning}
\bibinfo{author}{H.~Cheng}, \bibinfo{author}{Z.~Zhu}, \bibinfo{author}{X.~Li}, \bibinfo{author}{Y.~Gong}, \bibinfo{author}{X.~Sun}, \bibinfo{author}{Y.~Liu},
\newblock \bibinfo{title}{Learning with instance-dependent label noise: A sample sieve approach},
\newblock \bibinfo{journal}{arXiv preprint arXiv:2010.02347}  (\bibinfo{year}{2020}).
%Type = Article
\bibitem[{Zhu et~al.(2021)Zhu, Chen, Peng, Wang, and Jin}]{zhu2021hard}
\bibinfo{author}{C.~Zhu}, \bibinfo{author}{W.~Chen}, \bibinfo{author}{T.~Peng}, \bibinfo{author}{Y.~Wang}, \bibinfo{author}{M.~Jin},
\newblock \bibinfo{title}{Hard sample aware noise robust learning for histopathology image classification},
\newblock \bibinfo{journal}{IEEE transactions on medical imaging} \bibinfo{volume}{41} (\bibinfo{year}{2021}) \bibinfo{pages}{881--894}.
%Type = Article
\bibitem[{Zhang et~al.(2022)Zhang, Deng, Cui, Yin, Shi, and Wen}]{zhang2022model}
\bibinfo{author}{Y.~Zhang}, \bibinfo{author}{W.~Deng}, \bibinfo{author}{X.~Cui}, \bibinfo{author}{Y.~Yin}, \bibinfo{author}{H.~Shi}, \bibinfo{author}{D.~Wen},
\newblock \bibinfo{title}{Model and data agreement for learning with noisy labels},
\newblock \bibinfo{journal}{arXiv preprint arXiv:2212.01054}  (\bibinfo{year}{2022}).
%Type = Article
\bibitem[{Cordeiro et~al.(2023)Cordeiro, Sachdeva, Belagiannis, Reid, and Carneiro}]{cordeiro2023longremix}
\bibinfo{author}{F.~R. Cordeiro}, \bibinfo{author}{R.~Sachdeva}, \bibinfo{author}{V.~Belagiannis}, \bibinfo{author}{I.~Reid}, \bibinfo{author}{G.~Carneiro},
\newblock \bibinfo{title}{Longremix: Robust learning with high confidence samples in a noisy label environment},
\newblock \bibinfo{journal}{Pattern Recognition} \bibinfo{volume}{133} (\bibinfo{year}{2023}) \bibinfo{pages}{109013}.
%Type = Article
\bibitem[{Wen et~al.(2023)Wen, Xu, and Ying}]{wen2023jsmix}
\bibinfo{author}{Z.~Wen}, \bibinfo{author}{H.~Xu}, \bibinfo{author}{S.~Ying},
\newblock \bibinfo{title}{Jsmix: a holistic algorithm for learning with label noise},
\newblock \bibinfo{journal}{Neural Computing and Applications} \bibinfo{volume}{35} (\bibinfo{year}{2023}) \bibinfo{pages}{1519--1533}.
%Type = Article
\bibitem[{Li et~al.(2024)Li, Guo, Wang, and Xu}]{li2024103801}
\bibinfo{author}{Y.~Li}, \bibinfo{author}{Z.~Guo}, \bibinfo{author}{L.~Wang}, \bibinfo{author}{L.~Xu},
\newblock \bibinfo{title}{Tbc-mi : Suppressing noise labels by maximizing cleaning samples for robust image classification},
\newblock \bibinfo{journal}{Information Processing \& Management} \bibinfo{volume}{61} (\bibinfo{year}{2024}) \bibinfo{pages}{103801}. \URLprefix \url{https://www.sciencedirect.com/science/article/pii/S0306457324001602}. \DOIprefix\doi{https://doi.org/10.1016/j.ipm.2024.103801}.
%Type = Inproceedings
\bibitem[{Wan et~al.(2024)Wan, Wang, Xie, Li, Huang, and Chen}]{wan2024unlocking}
\bibinfo{author}{W.~Wan}, \bibinfo{author}{X.~Wang}, \bibinfo{author}{M.-K. Xie}, \bibinfo{author}{S.-Y. Li}, \bibinfo{author}{S.-J. Huang}, \bibinfo{author}{S.~Chen},
\newblock \bibinfo{title}{Unlocking the power of open set: A new perspective for open-set noisy label learning},
\newblock in: \bibinfo{booktitle}{Proceedings of the AAAI conference on artificial intelligence}, volume~\bibinfo{volume}{38}, \bibinfo{year}{2024}, pp. \bibinfo{pages}{15438--15446}.
%Type = Inproceedings
\bibitem[{Chen et~al.(2020{\natexlab{a}})Chen, Kornblith, Norouzi, and Hinton}]{chen2020simple}
\bibinfo{author}{T.~Chen}, \bibinfo{author}{S.~Kornblith}, \bibinfo{author}{M.~Norouzi}, \bibinfo{author}{G.~Hinton},
\newblock \bibinfo{title}{A simple framework for contrastive learning of visual representations},
\newblock in: \bibinfo{booktitle}{International conference on machine learning}, \bibinfo{organization}{PMLR}, \bibinfo{year}{2020}{\natexlab{a}}, pp. \bibinfo{pages}{1597--1607}.
%Type = Article
\bibitem[{Chen et~al.(2020{\natexlab{b}})Chen, Fan, Girshick, and He}]{chen2020improved}
\bibinfo{author}{X.~Chen}, \bibinfo{author}{H.~Fan}, \bibinfo{author}{R.~Girshick}, \bibinfo{author}{K.~He},
\newblock \bibinfo{title}{Improved baselines with momentum contrastive learning},
\newblock \bibinfo{journal}{arXiv preprint arXiv:2003.04297}  (\bibinfo{year}{2020}{\natexlab{b}}).
%Type = Inproceedings
\bibitem[{Ortego et~al.(2021)Ortego, Arazo, Albert, O'Connor, and McGuinness}]{ortego2021multi}
\bibinfo{author}{D.~Ortego}, \bibinfo{author}{E.~Arazo}, \bibinfo{author}{P.~Albert}, \bibinfo{author}{N.~E. O'Connor}, \bibinfo{author}{K.~McGuinness},
\newblock \bibinfo{title}{Multi-objective interpolation training for robustness to label noise},
\newblock in: \bibinfo{booktitle}{Proceedings of the IEEE/CVF Conference on Computer Vision and Pattern Recognition}, \bibinfo{year}{2021}, pp. \bibinfo{pages}{6606--6615}.
%Type = Inproceedings
\bibitem[{Yang et~al.(2021)Yang, Bisk, and Gao}]{yang2021taco}
\bibinfo{author}{J.~Yang}, \bibinfo{author}{Y.~Bisk}, \bibinfo{author}{J.~Gao},
\newblock \bibinfo{title}{Taco: Token-aware cascade contrastive learning for video-text alignment},
\newblock in: \bibinfo{booktitle}{Proceedings of the IEEE/CVF International Conference on Computer Vision}, \bibinfo{year}{2021}, pp. \bibinfo{pages}{11562--11572}.
%Type = Inproceedings
\bibitem[{Li et~al.(2021)Li, Xiong, and Hoi}]{li2021learning}
\bibinfo{author}{J.~Li}, \bibinfo{author}{C.~Xiong}, \bibinfo{author}{S.~C. Hoi},
\newblock \bibinfo{title}{Learning from noisy data with robust representation learning},
\newblock in: \bibinfo{booktitle}{Proceedings of the IEEE/CVF International Conference on Computer Vision}, \bibinfo{year}{2021}, pp. \bibinfo{pages}{9485--9494}.
%Type = Inproceedings
\bibitem[{Jiang et~al.(2020)Jiang, Huang, Liu, and Yang}]{jiang2020beyond}
\bibinfo{author}{L.~Jiang}, \bibinfo{author}{D.~Huang}, \bibinfo{author}{M.~Liu}, \bibinfo{author}{W.~Yang},
\newblock \bibinfo{title}{Beyond synthetic noise: Deep learning on controlled noisy labels},
\newblock in: \bibinfo{booktitle}{International conference on machine learning}, \bibinfo{organization}{PMLR}, \bibinfo{year}{2020}, pp. \bibinfo{pages}{4804--4815}.
%Type = Inproceedings
\bibitem[{Yan et~al.(2022)Yan, Luo, Xu, Deng, and Huang}]{yan2022noise}
\bibinfo{author}{J.~Yan}, \bibinfo{author}{L.~Luo}, \bibinfo{author}{C.~Xu}, \bibinfo{author}{C.~Deng}, \bibinfo{author}{H.~Huang},
\newblock \bibinfo{title}{Noise is also useful: Negative correlation-steered latent contrastive learning},
\newblock in: \bibinfo{booktitle}{Proceedings of the IEEE/CVF Conference on Computer Vision and Pattern Recognition}, \bibinfo{year}{2022}, pp. \bibinfo{pages}{31--40}.
%Type = Inproceedings
\bibitem[{Zhang et~al.(2017)Zhang, Bengio, Hardt, Recht, and Vinyals}]{DBLP:conf/iclr/ZhangBHRV17}
\bibinfo{author}{C.~Zhang}, \bibinfo{author}{S.~Bengio}, \bibinfo{author}{M.~Hardt}, \bibinfo{author}{B.~Recht}, \bibinfo{author}{O.~Vinyals},
\newblock \bibinfo{title}{Understanding deep learning requires rethinking generalization},
\newblock in: \bibinfo{booktitle}{5th International Conference on Learning Representations, {ICLR} 2017, Toulon, France, April 24-26, 2017, Conference Track Proceedings}, \bibinfo{publisher}{OpenReview.net}, \bibinfo{year}{2017}. \URLprefix \url{https://openreview.net/forum?id=Sy8gdB9xx}.
%Type = Article
\bibitem[{Sohn et~al.(2020)Sohn, Berthelot, Carlini, Zhang, Zhang, Raffel, Cubuk, Kurakin, and Li}]{sohn2020fixmatch}
\bibinfo{author}{K.~Sohn}, \bibinfo{author}{D.~Berthelot}, \bibinfo{author}{N.~Carlini}, \bibinfo{author}{Z.~Zhang}, \bibinfo{author}{H.~Zhang}, \bibinfo{author}{C.~A. Raffel}, \bibinfo{author}{E.~D. Cubuk}, \bibinfo{author}{A.~Kurakin}, \bibinfo{author}{C.-L. Li},
\newblock \bibinfo{title}{Fixmatch: Simplifying semi-supervised learning with consistency and confidence},
\newblock \bibinfo{journal}{Advances in neural information processing systems} \bibinfo{volume}{33} (\bibinfo{year}{2020}) \bibinfo{pages}{596--608}.
%Type = Inproceedings
\bibitem[{Zhao et~al.(2022)Zhao, Zhou, Wang, Shi, and Gao}]{zhao2022lassl}
\bibinfo{author}{Z.~Zhao}, \bibinfo{author}{L.~Zhou}, \bibinfo{author}{L.~Wang}, \bibinfo{author}{Y.~Shi}, \bibinfo{author}{Y.~Gao},
\newblock \bibinfo{title}{Lassl: Label-guided self-training for semi-supervised learning},
\newblock in: \bibinfo{booktitle}{Proceedings of the AAAI conference on artificial intelligence}, volume~\bibinfo{volume}{36}, \bibinfo{year}{2022}, pp. \bibinfo{pages}{9208--9216}.
%Type = Article
\bibitem[{Wang et~al.(2020)Wang, Chan, and Zeng}]{wang2020probabilistic}
\bibinfo{author}{L.~Wang}, \bibinfo{author}{R.~Chan}, \bibinfo{author}{T.~Zeng},
\newblock \bibinfo{title}{Probabilistic semi-supervised learning via sparse graph structure learning},
\newblock \bibinfo{journal}{IEEE transactions on neural networks and learning systems} \bibinfo{volume}{32} (\bibinfo{year}{2020}) \bibinfo{pages}{853--867}.
%Type = Article
\bibitem[{Gu et~al.(2016)Gu, Fan, and Meng}]{gu2016robust}
\bibinfo{author}{N.~Gu}, \bibinfo{author}{M.~Fan}, \bibinfo{author}{D.~Meng},
\newblock \bibinfo{title}{Robust semi-supervised classification for noisy labels based on self-paced learning},
\newblock \bibinfo{journal}{IEEE Signal Processing Letters} \bibinfo{volume}{23} (\bibinfo{year}{2016}) \bibinfo{pages}{1806--1810}.
%Type = Article
\bibitem[{Zhang et~al.(2022)Zhang, Xie, Li, Camargo, Song, Lu, Jeudy, Dreizin, Melhem, Wang et~al.}]{zhang2022improving}
\bibinfo{author}{L.~Zhang}, \bibinfo{author}{D.~Xie}, \bibinfo{author}{Y.~Li}, \bibinfo{author}{A.~Camargo}, \bibinfo{author}{D.~Song}, \bibinfo{author}{T.~Lu}, \bibinfo{author}{J.~Jeudy}, \bibinfo{author}{D.~Dreizin}, \bibinfo{author}{E.~R. Melhem}, \bibinfo{author}{Z.~Wang}, et~al.,
\newblock \bibinfo{title}{Improving sensitivity of arterial spin labeling perfusion mri in alzheimer's disease using transfer learning of deep learning-based asl denoising},
\newblock \bibinfo{journal}{Journal of Magnetic Resonance Imaging} \bibinfo{volume}{55} (\bibinfo{year}{2022}) \bibinfo{pages}{1710--1722}.
%Type = Article
\bibitem[{Krizhevsky et~al.(2009)Krizhevsky, Hinton et~al.}]{krizhevsky2009learning}
\bibinfo{author}{A.~Krizhevsky}, \bibinfo{author}{G.~Hinton}, et~al.,
\newblock \bibinfo{title}{Learning multiple layers of features from tiny images}  (\bibinfo{year}{2009}).
%Type = Inproceedings
\bibitem[{Bossard et~al.(2014)Bossard, Guillaumin, and Van~Gool}]{bossard2014food}
\bibinfo{author}{L.~Bossard}, \bibinfo{author}{M.~Guillaumin}, \bibinfo{author}{L.~Van~Gool},
\newblock \bibinfo{title}{Food-101--mining discriminative components with random forests},
\newblock in: \bibinfo{booktitle}{Computer Vision--ECCV 2014: 13th European Conference, Zurich, Switzerland, September 6-12, 2014, Proceedings, Part VI 13}, \bibinfo{organization}{Springer}, \bibinfo{year}{2014}, pp. \bibinfo{pages}{446--461}.
%Type = Article
\bibitem[{Malach and Shalev-Shwartz(2017)}]{malach2017decoupling}
\bibinfo{author}{E.~Malach}, \bibinfo{author}{S.~Shalev-Shwartz},
\newblock \bibinfo{title}{Decoupling" when to update" from" how to update"},
\newblock \bibinfo{journal}{Advances in neural information processing systems} \bibinfo{volume}{30} (\bibinfo{year}{2017}).
%Type = Inproceedings
\bibitem[{Wei et~al.(2020)Wei, Feng, Chen, and An}]{wei2020combating}
\bibinfo{author}{H.~Wei}, \bibinfo{author}{L.~Feng}, \bibinfo{author}{X.~Chen}, \bibinfo{author}{B.~An},
\newblock \bibinfo{title}{Combating noisy labels by agreement: A joint training method with co-regularization},
\newblock in: \bibinfo{booktitle}{Proceedings of the IEEE/CVF conference on computer vision and pattern recognition}, \bibinfo{year}{2020}, pp. \bibinfo{pages}{13726--13735}.
%Type = Inproceedings
\bibitem[{Song et~al.(2019)Song, Kim, and Lee}]{song2019selfie}
\bibinfo{author}{H.~Song}, \bibinfo{author}{M.~Kim}, \bibinfo{author}{J.-G. Lee},
\newblock \bibinfo{title}{Selfie: Refurbishing unclean samples for robust deep learning},
\newblock in: \bibinfo{booktitle}{International Conference on Machine Learning}, \bibinfo{organization}{PMLR}, \bibinfo{year}{2019}, pp. \bibinfo{pages}{5907--5915}.
%Type = Inproceedings
\bibitem[{Yi and Wu(2019)}]{yi2019probabilistic}
\bibinfo{author}{K.~Yi}, \bibinfo{author}{J.~Wu},
\newblock \bibinfo{title}{Probabilistic end-to-end noise correction for learning with noisy labels},
\newblock in: \bibinfo{booktitle}{Proceedings of the IEEE/CVF conference on computer vision and pattern recognition}, \bibinfo{year}{2019}, pp. \bibinfo{pages}{7017--7025}.
%Type = Inproceedings
\bibitem[{Peng et~al.(2020)Peng, Wang, Zeng, Li, Yang, and Qiao}]{peng2020suppressing}
\bibinfo{author}{X.~Peng}, \bibinfo{author}{K.~Wang}, \bibinfo{author}{Z.~Zeng}, \bibinfo{author}{Q.~Li}, \bibinfo{author}{J.~Yang}, \bibinfo{author}{Y.~Qiao},
\newblock \bibinfo{title}{Suppressing mislabeled data via grouping and self-attention},
\newblock in: \bibinfo{booktitle}{Computer Vision--ECCV 2020: 16th European Conference, Glasgow, UK, August 23--28, 2020, Proceedings, Part XVI 16}, \bibinfo{organization}{Springer}, \bibinfo{year}{2020}, pp. \bibinfo{pages}{786--802}.
%Type = Article
\bibitem[{Huang et~al.(2020)Huang, Zhang, and Zhang}]{huang2020self}
\bibinfo{author}{L.~Huang}, \bibinfo{author}{C.~Zhang}, \bibinfo{author}{H.~Zhang},
\newblock \bibinfo{title}{Self-adaptive training: beyond empirical risk minimization},
\newblock \bibinfo{journal}{Advances in neural information processing systems} \bibinfo{volume}{33} (\bibinfo{year}{2020}) \bibinfo{pages}{19365--19376}.
%Type = Article
\bibitem[{Zhang et~al.(2021)Zhang, Zheng, Wu, Goswami, and Chen}]{zhang2021learning}
\bibinfo{author}{Y.~Zhang}, \bibinfo{author}{S.~Zheng}, \bibinfo{author}{P.~Wu}, \bibinfo{author}{M.~Goswami}, \bibinfo{author}{C.~Chen},
\newblock \bibinfo{title}{Learning with feature-dependent label noise: A progressive approach},
\newblock \bibinfo{journal}{arXiv preprint arXiv:2103.07756}  (\bibinfo{year}{2021}).
%Type = Inproceedings
\bibitem[{Han et~al.(2019)Han, Luo, and Wang}]{han2019deep}
\bibinfo{author}{J.~Han}, \bibinfo{author}{P.~Luo}, \bibinfo{author}{X.~Wang},
\newblock \bibinfo{title}{Deep self-learning from noisy labels},
\newblock in: \bibinfo{booktitle}{Proceedings of the IEEE/CVF international conference on computer vision}, \bibinfo{year}{2019}, pp. \bibinfo{pages}{5138--5147}.

\end{thebibliography}
% Use the style of numbering in square brackets.
% If nothing is used, default style will be taken.
%\begin{enumerate}[a)]
%\item 
%\item 
%\item 
%\end{enumerate}  

%\begin{itemize}
%\item 
%\item 
%\item 
%\end{itemize}  

%\begin{description}
%\item[]
%\item[] 
%\item[] 
%\end{description}  

% Figure
% \begin{figure}[<options>]
% 	\centering
% 		\includegraphics[<options>]{}
% 	  \caption{}\label{fig1}
% \end{figure}

% \begin{table}[<options>]
% \begin{tabular*}{\tblwidth}{@{}LL@{}}
% \toprule
% \midrule
%  & \\
%  & \\
%  & \\
%  & \\
% \bottomrule
% \end{tabular*}
% \end{table}

% Uncomment and use as the case may be
%\begin{theorem} 
%\end{theorem}

% Uncomment and use as the case may be
%\begin{lemma} 
%\end{lemma}



%\bibliographystyle{cas-model2-names}
%\biboptions{authoryears}
%\bibliographystyle{mybibfile}
%\bibliography{cas-refs}



\end{document}